\documentclass[
  bibliography=totoc,
  DIV=12,
  fleqn,
]{scrartcl}
\usepackage{mathtools}
\usepackage{empheq}
\usepackage{todonotes}
\KOMAoptions{numbers=noenddot}
\makeatletter
\renewcommand*\@seccntformat[1]{%
  \ifcsname the#1\endcsname
    \csname the#1\endcsname
    \ifstrequal{#1}{section}{.\ }{%
      \ifboolexpr{
        test {\ifstrequal{#1}{subsection}}
        or
        test {\ifstrequal{#1}{subsubsection}}
      }{%
        \ %
      }{%
        .\ %
      }%
    }%
  \fi}
\makeatother
\setcapindent{0pt}
\setkomafont{captionlabel}{\bfseries}
\usepackage{hyperref}
\usepackage{ragged2e}
\usepackage{makecell}
\usepackage{float}
\usepackage{multirow}
\usepackage{rotating}
\usepackage{adjustbox}
\usepackage{amsthm}
\usepackage{thmtools}
\usepackage{acro}
\usepackage{makecell}
\usepackage{makeidx}
\usepackage{xspace}
\usepackage{listings}
\usepackage{cite}
\usepackage{xcolor}
\usepackage{tikz}
\usetikzlibrary{hobby}
\usepackage[compatibility]{circuitikz}
\usepackage{longtable}
\usetikzlibrary {circuits.ee.IEC}
\usepackage{pgfplots}
\usepackage{subfig}
\usepgfplotslibrary{groupplots,dateplot}
\usetikzlibrary{automata,patterns,shapes.arrows}
\usetikzlibrary{arrows}
\usetikzlibrary{positioning}
\usetikzlibrary{arrows.meta}
\usetikzlibrary{shapes}
\usetikzlibrary{graphs}
\usetikzlibrary{quotes}
\usepackage{tikz-3dplot}
\usepackage{listofitems}
\usetikzlibrary {shapes.arrows}
\pgfplotsset{compat=newest}
\usepackage{rotating}
\usepackage{adjustbox}
\usepackage{amsmath,amssymb,amsfonts}
\usepackage{mathtools}
\usepackage{algorithm2e}
\usepackage{algorithmic}
\usepackage{afterpage}
\usepackage{graphicx}
\usepackage{caption}
\usepackage{subcaption}
\usepackage{textcomp}
\usepackage{xcolor}
\def\BibTeX{{\rm B\kern-.05em{\sc i\kern-.025em b}\kern-.08em
    T\kern-.1667em\lower.7ex\hbox{E}\kern-.125emX}}
\definecolor{TealBlue}{RGB}{0,128,128}
\usepackage[utf8]{inputenc}
\usepackage[T1]{fontenc}
\usepackage{nat_preamble}
\usepackage{acro}
\DeclareAcronym{RV}{
short = RV,
long  = \emph{Random Variable},
tag   = abbrev}
\DeclareAcronym{CF}{
short = CF,
long  = \emph{Circular Factory},
tag   = abbrev}
\DeclareAcronym{GMM}{
short = GMM,
long  = \emph{Gaussian Mixture Model},
tag   = abbrev}
\DeclareAcronym{PDF}{
short = PDF,
long  = \emph{Probability Density Function},
tag   = abbrev}
\DeclareAcronym{DOB}{
short = DoB,
long  = \emph{Degree of Belief},
tag   = abbrev}
\DeclareAcronym{EM}{
short = EM,
long  = \emph{Expectation Maximization},
tag   = abbrev}
\DeclareAcronym{BIC}{
short = BIC,
long  = \emph{Bayesian Information Criterion},
tag   = abbrev}
\DeclareAcronym{AIC}{
short = AIC,
long  = \emph{Akaike Information Criterion},
tag   = abbrev}

\newtheorem{corollary}{Corollary}[section]

\newtheorem{proposition}{Proposition}[section]
\newtheorem{definition}{Definition}[section]

\newtheorem{remark}{Remark}[section]
\setkomafont{disposition}{\normalfont\bfseries}
\usepackage{authblk}
\usepackage{cleveref}
\title{
  Comprehensive Description of Uncertainty in Measurement for Representation and Propagation with Scalable Precision
}
\author[1,2]{Ali Darijani}
\author[1,2]{Jürgen Beyerer}
\author[1,2]{Zahra Sadat Hajseyed Nasrollah}
\author[3]{Luisa Hoffmann}
\author[3]{Michael Heizmann}
\affil[1]{Fraunhofer IOSB}
\affil[2]{KIT, IES}
\affil[3]{KIT, IIIT}
\date{\today}
\begin{document}
\maketitle
\begin{abstract}
Probability theory has become the predominant framework for quantifying uncertainty across
scientific and engineering disciplines, with a particular focus on measurement and control systems. However, the widespread reliance on simple Gaussian assumptions—particularly in
control theory, manufacturing, and measurement systems—can result in incomplete representations and
multistage lossy approximations of complex phenomena, including inaccurate propagation of uncertainty through multi stage processes.

This work proposes a comprehensive yet computationally tractable framework for representing and propagating quantitative attributes arising in measurement systems
using \acp{PDF}. Recognizing the constraints imposed by finite memory in software systems,
we advocate for the use of \acp{GMM}, a principled extension of the familiar Gaussian framework, as they are universal approximators of \acp{PDF} whose complexity can be tuned to trade off approximation accuracy against memory and computation. From both mathematical and computational perspectives, \acp{GMM} enable high performance and, in many cases,
closed form solutions of essential operations in control and measurement.

The paper presents practical applications within manufacturing and measurement contexts especially circular factory, demonstrating how the \acp{GMM} framework
supports accurate representation and propagation of measurement uncertainty and offers improved accuracy--compared to the traditional Gaussian framework--while keeping the computations tractable.

\end{abstract}
\section{Introduction}
In this paper, we present a consistent, generic framework to efficiently represent and propagate measurement uncertainties 
in industrial systems with scalable accuracy. When measuring quantitative attributes, it is not sufficient to 
just report a numerical value for the measurement result. Only together with a quantitative description of 
the measurement uncertainty about the measured attribute value does the result become usable. According 
to the state of the art in metrology, uncertainty is described on the basis of a number with the 
semantics of a standard deviation or multiples thereof~\cite{JCGM2008}. To 
reduce stochastic errors, the measurement result usually emanates from an averaging calculation on 
the observed data. From a probabilistic point of view, with the measurement result and the uncertainty in 
form of a mean value and a standard deviation, respectively, a Gaussian distribution for perturbations is 
implicitly assumed regarding the formation of observable data based on the underlying measurand.
In the Guide to the Expression of Uncertainty in Measurement~\cite{JCGM2008}, however, 
this methodology is used not only to describe aleatory uncertainties (due to randomness), 
but also to represent epistemic uncertainties (due to lack of knowledge), e.g. to quantify 
systematic but unknown errors. The interpretation of probability used then corresponds to a \ac{DOB}, 
which allows non-random uncertainties and expert knowledge to be represented and treated mathematically 
consistently using the calculus of probability. The \ac{DOB} is the more general interpretation of probability 
that is compatible with Kolmogorov's axioms and allows a unified quantitative description of aleatory and epistemic uncertainties. 

The reason why in this paper, as well as by the metrological community, uncertainties are 
regularly described by means of probabilities and alternative calculus for describing uncertainty 
do not play a role is explained in \Cref{sec:philosophical_aspects} and \Cref{sec:mathematical_aspects}. In practice, there are many cases where a Gaussian 
assumption (implicitly or explicitly) is not adequate; see \Cref{sec:gaussian_insufficiency}. In particular, 
in the propagation of measurement results and their uncertainties through multi-stage systems, errors accumulate 
along the calculations due to the simple assumption of Gaussian distributions. Consequently, the uncertainty 
would then ideally have to be described by means of a complete \ac{PDF}, but this is generally mathematically 
very challenging in practice; see \Cref{sec:nonlinear_transformation}. 
In this paper, we propose to describe \acp{PDF} using \acp{GMM}. These are 
universal approximators for \acp{PDF}~\cite{Mazja2014}\cite{Lu2017}\cite{Huber2015}, where the precision of the approximation can be specifically 
adjusted by the number of components of the \ac{GMM}; see \Cref{sec:gmm_universal_approximator}. This leads 
to a finitely dimensional, usually even low-dimensional description of the \acp{PDF} using the parameter tuple \ac{GMM}.

We even suggest that all uncertain quantitative attributes that play a role in a measurement process should be 
described as \ac{PDF} in the form of \ac{GMM}, i.e. \ac{GMM} should be regarded as the ``native data type'' of such quantities. 
Many practically relevant operations, such as the addition of quantities, Bayesian fusion of different information with respect 
to a quantity of interest, the mixing of similar quantities from different sources, the focusing on individual 
quantities from a set of different quantities, and the exploitation of observed quantities for inference with respect 
to unobserved quantities, can be performed by algebraic operations on the parameter tuples of the \acp{GMM} with simple 
formulas, very efficiently and accurately; see \Cref{tab:gm_math}. For operations between quantities 
for which an algebraic calculation of the parameters of the \ac{GMM} of the result is not possible, 
a mapping to a description using \acp{GMM} can be accomplished by efficient Monte Carlo sampling and subsequent 
fitting of a \ac{GMM} using the \ac{EM} algorithm; see \Cref{sec:gmm_sampling} and \Cref{sec:gmm_fitting}.
Explanatory examples for the operations in \Cref{tab:gm_math} are discussed in \Cref{sec:examples}. 

In addition to the uniform description of uncertain quantities using \acp{GMM}, measuring instruments and systems 
can also be described in a scalable and probabilistic manner also by using \acp{GMM}; see \Cref{sec:measurement_system}. Based 
on this representation, the change in the uncertainty of a quantity of interest due to a 
measurement operation can also be calculated purely on the basis of algebraic operations on the parameter tuples of \acp{GMM}. From a 
Bayesian point of view, prior knowledge about the quantity of interest can be embodied as \ac{GMM}. The observation of 
measurement data converts the \ac{GMM} of the measurement system into the likelihood function in the form of a \ac{GMM}, which 
can then be fused algebraically using Bayes' theorem with the prior knowledge to achieve the posterior \ac{PDF} 
in the form of a \ac{GMM} as the final measurement result. Hence, measurement processes can be described by means of 
algebraic operations on the parameter tuples of \acp{GMM}. 

At first glance, the formula apparatus in \Cref{tab:gm_math}, together 
with the mathematical details listed in \Cref{sec:appendix}, could be a deterrent for practical use. In 
terms of software, the representations and operations can be encapsulated in detail and the user is shielded from them. The user can 
handle the quantities in the usual manner and only in the background the quantities are treated and calculated as \acp{GMM}. In multi-stage 
systems, this can be used to calculate more accurately internally, whereby a reversal to the usual description 
by means of a mean value and a standard deviation is possible at any time and can be 
calculated algebraically from the \ac{GMM} without any problems; see \Cref{sec:gaussian_fallback}. 

\section{Mastering Uncertainty within the Circular Factory}

In the Circular Factory, full-fledged products are made mixed from new and used parts. 
Since used parts have different histories and usage durations behind them, it does not 
make sense to want to represent the ensemble of a certain component type as a statistical 
population with only one single \ac{PDF} with regard to its quality-relevant and 
function-relevant properties (attributes). Rather, an individual probabilistic description 
of the relevant properties is required for each component instance. For continuous attributes, 
this can be achieved with scalable accuracy and efficient handling by means of 
component-specific \acp{GMM}. These \acp{GMM} are determined in such a way that they reflect 
the individual level of information about the attributes of a specific component, including 
the associated uncertainties and correlations.

In the Circular Factory probability is consistently interpreted as \ac{DOB}, so that 
probabilities based on statistical evidence as well as subjective expert knowledge 
and opinions can be described and handled uniformly. Each instance of a component is 
therefore linked to an individual data structure that reflects the current level of 
information about the quantitative attributes relevant to function and quality in 
the form of \ac{GMM} parameter tuples.
\\If further information about a component instance is to be added, e.g. through a measurement, 
the stored \ac{GMM} is understood as the prior distribution and the additional information 
as a likelihood. Both are fused in the Bayesian sense to the posterior distribution in the form 
of an updated \ac{GMM}.

Overall, this approach allows all relevant component attributes, 
including their aleatoric as well as epistemic uncertainties, to be accurately 
represented in an instance-individually adjustable manner. In addition, 
all important operations regarding uncertainties and their propagation can 
be carried out very efficiently in the Circular Factory, mainly by means of 
closed calculations and consistently based on \acp{GMM}.

\section{Uncertainty Representation and the Probability Framework}
Probabilities provide a consistent and powerful framework for
representing uncertainty and the subsequent decision making
and reasoning under it. Its adaptation by both practitioners
and theoreticians comes from its historical success and the math that powers it.

\subsection{Philosophical Aspects and Historical Success}
\label{sec:philosophical_aspects}
In modern era much of the philosophical aspect success of
the probability is due to \cite{Finetti2017}.
Finetti showed by a specific reward function based on a decision
that is to be made under uncertainty, one must adhere to
probability to maximize their reward. Later Lindley
relaxed the assumptions on the reward function
which ascertained the place of probability in
uncertainty quantification \cite{Lindley1982}.
Interestingly, he also challenged those proposing
other nonprobabilistic frameworks--mainly Lotfi Zadeh
and Arthur Dempster-- arguing that
everything that can be done within their frameworks can be done
better using probability \cite{Lindley1987}. 

Within probability there
are different interpretations--mainly classical \cite{Laplace1820}, frequentist \cite{Venn1888}, Bayesian \cite{Bayes1763}--which lead to different formulations and all having their pros and cons. Classical interpretation is fairly limited and can only answer simple queries such as ``What is the probability that the outcome of a dice roll is an odd number
given that we have a fair dice?'' which requires idealization of 
the experiment as hinted by the assumption that the dice is fair which 
is also known as the equally likely assumption and quite strong 
symmetry. The frequentist interpretation, also known as 
the asymptotic interpretation, defines probability as the 
long run relative frequency of an event in repeated identical 
trials which is limited as one cannot hope to have identical 
trials even for two experiments let alone quite a large number 
of them. 

The Bayesian interpretation or also known as the 
subjective interpretation formulates the \ac{DOB} one might 
have about the outcome of an experiment as a \ac{PDF} and would 
update the \ac{PDF} given new experimental results or new data 
through the Bayes' theorem. Two main advantages of the Bayesian 
interpretation are that it can draw conclusion based on zero 
number of data points and it is suitable for online learning and 
online inference. The challenge is that the decisions drawn 
from the Bayesian interpretation when having small amount of 
data depends on the prior knowledge or the expert knowledge, 
formulated as a \ac{DOB} \ac{PDF}, and can only become better by 
having a better expert or by acquiring new data. In the context of 
scientific machine learning, measurement systems, and physical 
systems there is a strong inclination to use the Bayesian interpretation 
due to lack of data and the expensive cost of performing experiments.

\subsection{Mathematical Aspects}
\label{sec:mathematical_aspects}
Probability theory does not only have historical success in solving real
world problems, but also has solid mathematical foundations. It
possesses a well defined calculus--Kolmogorov's formulation
\cite{Kolmogoroff1933}--based on measure theory and by extension analysis \cite{Tao2011}. All the interpretations described in \Cref{sec:philosophical_aspects} 
are compatible with Kolmogorov's axiomatic foundations of probability theory and with the corresponding probability calculus~\cite{Beyerer2024}.
 This foundation forms the basis of modern
statistical inference and probabilistic machine learning.

\section{Uncertainty in Measurement, the Gaussian Framework and its Insufficiency}
Measurement uncertainty~\cite{Gerlach2024} is usually being handled using the Gaussian
framework. While the Gaussian framework has been a powerful
tool it has inherent limitations in nonlinear systems and multimodal situations. 
According to \cite{JCGM2008a}, measurement uncertainty is 
determined, represented, and propagated based on standard deviations 
(root of variance). Restricting uncertainty to the notion of standard
deviation implicitly induces a Gaussian representation by virtue of the principle of maximum entropy \cite{Cover2006}.

\subsection{The Gaussian Framework}
Under the Gaussian assumption, measurement uncertainty is represented by a normal distribution characterized 
by its mean and standard deviation. This model implies that most measurements cluster around the mean, with 
decreasing probabilities for values farther away. Consequently, the uncertainty can be quantified and visualized 
effectively using confidence intervals.
 
\subsubsection{The Advantages of the Gaussian Framework}\label{sec:gaussian_advantages}
Assuming a Gaussian \ac{PDF} offers theoretical and computational benefits. 
Given an expected value and a variance a Gaussian is the representation that 
stays impartial or uncertain the most \cite{Cover2006} which for example in 
the context of reliability models means they become safer to 
use as they would overestimate the danger rather than underestimate it. 
If only the expected value $m$ and the variance $\sigma^2$ are known
about a \ac{RV} $X$, a Gaussian $\mathcal{N}_X(x|m,\sigma^2)$ is 
that \ac{PDF} with these parameters with the maximum entropy \cite{Cover2006}. Therefore, 
describing measurement results based on a mean value and measurement uncertainty by
 a variance implicitly assumes a Gaussian distribution. Gaussian formulation has many nice properties \cite{Tong1990}\cite{Rencher2012}\cite{Genz2009} that make decision making and inference faster and more robust.

\subsection{The Insufficiency of the Gaussian Framework}\label{sec:gaussian_insufficiency}
Gaussian models assume symmetry, light tails, linearity, and additive, homoscedastic noise. Real world data often exhibit skewed bahavior, heavy tails, multimodality, constraints, discontinuities, outliers, and nonlinear dynamics, violating these assumptions and therefore leading to miscalculation of
risk and other values of interest, if Gaussian distributions are assumed.

\subsubsection{Inherently Non Gaussian Physics}
While at the macro scale and low velocity most of the physical 
phenomena that exhibit randomness are modeled using the Gaussian 
framework, in smaller scales and high velocity the Gaussian 
framework fails. To name a few Relativistic Breit-Wigner 
distribution \cite{Breit1936}, quantum mechanical operators \cite{Schwabl2007}, 
electromagnetic operators \cite{Jackson2021} cannot be handled using the 
Gaussian framework.

\subsubsection{Nonlinear Transformation}
\label{sec:nonlinear_transformation}
Nonlinear transformation of inputs that are distributed according to a 
Gaussian in general are non Gaussian. The sensitivity analysis which 
is based on the first order Taylor series is insufficient to 
propagate the uncertainty forward and a loss of information will occur. The 
canonical example is that if two variables are normally distributed 
around zero, their ratio is Cauchy distributed and as the expectation value 
and the variance are not defined for the Cauchy distribution, therefore it is impossible to approximate it using a Gaussian distributed variable.

\subsubsection{Mixture of Sources}
If we have different sources for observing values for one and the same quantity $X$, and if the sources are independent and active 
with different probabilities, a common \ac{PDF} for the pool of these sources can be 
established in the form of the convex linear combination of the \acp{PDF} of the individual sources. In general, 
a linear combination of Gaussian \acp{PDF} does not result in a Gaussian \ac{PDF}, thus, the Gaussian framework is not sufficient anymore in this case~\cite{Chen2023}\cite{Titterington1995}.

\subsubsection{Constraint Handling}
As an example, modeling a positive physical quantity like length 
within the Gaussian framework might cause problems when 
propagated using expansion methods or Monte Carlo methods if 
the mean is close to zero and the standard deviation is quite large. In general hard constraints cannot be directly incorporated into the Gaussian framework due to the infinite support of the domain of the Gaussian variables.

\subsubsection{Abstract Non Gaussian Distributions}
Beyond physical quantities, there exist abstract 
non Gaussian probability distributions that have 
diverse applications in different disciplines. These 
distributions play a crucial role in various fields, 
offering tools for modeling complex phenomena that do 
not conform to traditional Gaussian assumptions. Non Gaussian phenomena in finance \cite{Jondeau2007} is probably the one that cannot be categorized under any physical phenomena and still exhibit non Gaussian behaviour.

\section{Gaussian Mixture Model and Its Advantages}
\acp{GMM} are the natural extension of the Gaussian framework which are simply the weighted sum of finitely many Gaussian \acp{PDF}. A list of nice properties of \acp{GMM} is presented in the \Cref{tab:gm_math}.
Doing statistical inference under the \ac{GMM} assumption provides mathematical and computational benefits. It is worth giving an exact mathematical definition of a \ac{GMM} before discussing its advantages as it provides a solid ground for the subsequent discussions.

\begin{definition}\label{def:gmm}
  Let
\begin{equation}
  \begin{split}
    &d, K                     \in \mathbb{N}\setminus \{0,+\infty\}\\
    &x \in  \mathbb{R}^d,m_i \in \mathbb{R}^d \text{ for } {i=1,\dots,K} \\
  &\{\Sigma_i\}_{i=1,\dots,K} \subset \{A \in \mathbb{R}^{d \times d}\colon x^\top \boldsymbol{\cdot}A\boldsymbol{\cdot}x>0 \text{ for } x \neq 0\}\\
    &\pi_i \in \mathbb{R}_+  \text{ for }{i=1,\dots,K}\land \sum_{i=1}^K \pi_i = 1,
  \end{split}
\end{equation}

then the mapping

\begin{equation}
  \mathcal{G}_X(x|K,\{\pi_i,m_i,\Sigma_i\}_{i=1,\dots,K}) \coloneq \sum_{i=1}^K\pi_i \mathcal{N}_X(x|m_i,\Sigma_i),
\end{equation}

is called a Gaussian Mixture or a \ac{GMM}.

\end{definition}
\begin{remark}[Notation]
  If we encapsulate all the parameters required to define a \ac{GMM} then:
  \begin{equation}
    \begin{split}
    &g                 \coloneq (K,\{\pi_i,m_i,\Sigma_i\}_{i=1,\dots,K})\\
    &\mathcal{G}_X(x|g)\coloneq \mathcal{G}_X(x|K,\{\pi_i,m_i,\Sigma_i\}_{i=1,\dots,K}).
    \end{split}
  \end{equation}
\end{remark}

\begin{remark}[Notation]
  When the Random Variable $X$ is Gaussian Mixture distributed, we write $X \sim \mathcal{G}_X(x|g)$.
\end{remark}

Doing statistical inference under the \ac{GMM} assumption provides mathematical and computational benefits.

\subsection{Mathematical Arguments}

\subsubsection{Universal Density Approximation}
\label{sec:gmm_universal_approximator}
\ac{GMM} is a universal \ac{PDF} approximator.
For a target \ac{PDF} $p_X(x)$ having enough regularity and an
appropriate dissimilarity measure $\mathrm{d}(\boldsymbol{\cdot},\boldsymbol{\cdot})$, 
a series of \acp{GMM} $\mathcal{G}_X(x|K,\{\pi_i,m_i,\Sigma_i\}_{i=1,\dots,K})$ exist such that 
$\mathcal{G}_X(x|g)$ converges to $p_X(x)$: $\lim_{K \to +\infty} \mathrm{d}(p_X(x),\mathcal{G}_X(x|g))$. i.e., 
by increasing the number of components, a \ac{GMM} can approximate the target \ac{PDF} $p_X(x)$ with any required precision~\cite{Lu2017}\cite{Mazja2014}.

\subsubsection{Fitting a Gaussian Mixture Model}\label{sec:gmm_fitting}
One of the standard ways to fit a \ac{GMM} to data is through
an iterative algorithm called \ac{EM}~\cite{Dempster1977}. \ac{EM} 
algorithm is highly efficient in the case of a \ac{GMM} as it
provides closed form updates at each iteration~\cite{Bishop2019}. The 
convergence is also guaranteed due to the celebrated Banach fixed point 
theorem.

\subsubsection{Closure Under Important Operations}
\label{sec:gmm_closure}
\acp{GMM} are closed under fundamental operations
that are of interest in statistics.
Affine transformations preserve mixture form,
allowing propagation through linear mappings without
approximation. Marginals and conditionals of a joint
\ac{GMM} remain a \ac{GMM}. Exact formulas in \Cref{prop:gmm_marginalization} and \Cref{prop:gmm_conditioning}
enable exact uncertainty quantification of the
variables. The product of two \acp{GMM} is
an unnormalized \ac{GMM} which
enables closed form calculation of some
dissimilarity measures and more importantly the fusion based on Bayes' theorem. Convolution of two \ac{GMM} is again a \ac{GMM} with the explicit formula as seen in \Cref{prop:gaussian_convolution} in \Cref{sec:appendix}.

\subsubsection{High Quality Generative Property}
\label{sec:gmm_sampling}
Sampling from a \ac{GMM} is a straightforward process.
Using composition method \cite{Robert2004} and the Box-Muller \cite{Robert2004} transform high quality 
samples can be drawn from a \ac{GMM}. The composition method selects the component at random and then a 
sample is drawn from the corresponding Gaussian making a two step sequential process requiring $d+1$ \ac{RV}s. This 
makes \ac{GMM} an attractive model as a generative model which has applications
in data augmentation, data completion and data imputation.

\subsubsection{Finite Dimensional}
Statistical inference over general probability distributions
is inherently an infinite-dimensional problem, as it involves
reasoning over function spaces with no finite representation.
Gaussian mixture framework provides a way to represent distributions
with a finite set of parameters. Having a finite set of parameters by
itself makes it easier to do inference since calculations in infinite
dimensional spaces are inherently more difficult and theoretically
involved~\cite{Clarke2013}.

\subsubsection{High Dimensional Integrals}
Many statistical properties and operations--such
as marginalization, conditioning, expectations,
and entropy--are fundamentally defined in terms
of integrals of functions of \acp{PDF}~\cite{Klenke2020} over usually infinite domains. In
high dimensional spaces, these integrals are often
intractable for general distributions and suffer from the curse of high dimensionality \cite{Vershynin2018}\cite{Wainwright2019}. However,
when working within the Gaussian mixture framework,
a remarkable number of these operations including but not limited to marginalization, conditioning, expectation, and variance
 admit closed
form or highly efficient analytic approximations.

\subsubsection{Gaussian Fallback}
\label{sec:gaussian_fallback}
\acp{GMM} provide a powerful yet flexible
framework for representing complex distributions
while preserving compatibility with the Gaussian framework.
When configured with a single component, the Gaussian mixture framework
reduces to the Gaussian framework, allowing users to opt for
simplicity when needed. A Gaussian distribution can approximate a
Gaussian mixture up to the second central moment.
The expectation and the variance of a mixture can be calculated
in closed form \cite{Barber2012} and used as parameters of an equivalent Gaussian.
This enables seamless fallback to the familiar Gaussian framework. The expectation and the variance of the correspondent Gaussian are given in the following propositions.

\begin{proposition}[\ac{GMM} Expectation]
\label{prop:gmm_expectation}
\begin{equation}
\label{eq:gmm_expectation}
    \mathbb{E}(X \sim \mathcal{G}_X(x|g))=\sum_{i=1}^{K}\pi_i m_i\\
\end{equation}
\end{proposition}

\begin{proposition}[\ac{GMM} Covariance Matrix]
\label{prop:gmm_variance}
\begin{equation}
\label{eq:gmm_variance}
    \mathbb{V}(X \sim \mathcal{G}_X(x|g))=\sum_{i=1}^{K}\pi_i(\Sigma_i+ {m_i \boldsymbol{\cdot} {m_i}^\top})-{(\sum_{i=1}^{K}\pi_i m_i)\boldsymbol{\cdot} (\sum_{i=1}^{K}\pi_i m_i)^\top}
\end{equation}
\end{proposition}

\subsection{Computational Arguments}

\subsubsection{Parallelizability of Gaussian Mixture Fitting}
\ac{EM} \cite{Dempster1977} is often used to fit a \ac{GMM}
to data which innately is parallelizable. The $\mathrm{E}$-step is
what the high performance computing community calls ``embarrassingly parallelizable''~\cite{Herlihy2012}
which roughly means that if you have $N$ computing resources the runtime is going
to be $1/N$ times the runtime it would taken using only $1$ computing resource.
The $\mathrm{M}$-step is also embarrassingly parallelizable as it
only uses sum reduction. This results in a linear speedup within the iteration step.
Streaming \ac{EM} enable online updates making it
invaluable for real time scenarios. Implementations fit well with distributed computing
which in turn result in scaling being feasible.

\subsubsection{Small Storage Footprint}
\label{sec:small_footprint}
On a computer one needs $2$ integers, the number of components $K$
and the dimension $d$, and $K-1+Kd+1/2Kd(d+1)$~\cite{Beyerer2024} floating point variables
to fully define a Gaussian mixture. This allows for a compact and efficient representation and access of the model both at fitting and inference time. Due to the additive nature of the Gaussian mixture it is even possible to split the model and store it on different nodes for again both at fitting and inference time.

\subsection{Closed Form Calculations}
By representing attributes and their uncertainties by \acp{GMM}, 
many practical relevant operations with attributes can be calculated 
algebraically on the parameter tuples $g$. \Cref{tab:gm_math} lists 
those operations together with the closed form formulas. 

E.g. if two attributed with the same physical meaning are added and if attributes 
are treated as stochastically independent \acp{RV}, the \acp{PDF} in
 form of \acp{GMM} are convolved. This results again in a \ac{GMM}, 
the parameters of which can be calculated in closed form from the parameters 
of the two \acp{GMM}; see \Cref{eq:tab:convolution} in \Cref{tab:gm_math}.

{\footnotesize
\clearpage
\thispagestyle{empty} 
\begin{longtable}{|>{\centering\arraybackslash}m{3.5cm}|>{\centering\arraybackslash}m{4.6cm}|>{\centering\arraybackslash}m{0.5cm}|>{\centering\arraybackslash}m{7.5cm}|}
\caption{This table provides a compact overview of the use of \acp{GMM} for the representation and combination 
of uncertain quantitative attributes and their \acp{PDF}. Column 1 lists the most important operations for the 
practical handling of attributes in real-world applications. Column 2 explains the associated mathematical 
meaning. The computation type (CT) in column 3 indicates whether the calculations can be accomplished algebraically 
on the parameter tuples of the \acp{GMM} (CF: Closed Form) or whether a numerical calculation (N) is required. 
Finally, in column 4, the closed formulas for the algebraic operations on the \ac{GMM} parameters are given 
compactly. $X \perp Y$ means that the \acp{RV} $X$ and $Y$ are stochastically independent. Additional details to these formulas are presented in \Cref{sec:appendix}. For each row of the table there 
is an explanatory concrete example in \Cref{sec:examples}.}
\label{tab:gm_math} \\
\hline
\textbf{Practical Meaning}& \textbf{Mathematical Expression} & \textbf{CT} & \textbf{Formulas/Algorithms} \\
\hline
\endfirsthead

\hline
\textbf{Practical Meaning} & \textbf{Mathematical Expression} & \textbf{CT} & \textbf{Formulas/Algorithms} \\

\hline
\endhead

\hline
\multicolumn{4}{r}{\textit{Continued on next page}} \\
\endfoot

\hline
\endlastfoot
\parbox[c][1.5cm][c]{7.5cm}{
Traditional \\measurement \\result; value $m$
} &
\parbox[c][1.5cm][c]{7.5cm}{
Expectation of \ac{RV}s; see \\\Cref{eq:gmm_expectation} in \Cref{prop:gmm_expectation}
} &
\parbox[c][1.5cm][c]{7.5cm}{
CF\textsuperscript{}
} &
\parbox[c][1.5cm][c]{7.5cm}{
    \begin{equation}
  \begin{aligned}
  &m = \sum_{i=1}^{K}\pi_i m_i
  \end{aligned}
	    \label{eq:expectation_in_table}
\end{equation}
} \\
\hline
\parbox[c][2.5cm][c]{6.5cm}{
Traditional \\measurement \\uncertainty $\pm k\sigma$; \\$\sigma^2$ = variance
} &
\parbox[c][2.5cm][c]{6.5cm}{
Variance of \acp{RV}; see \Cref{eq:gmm_variance} \\in \Cref{prop:gmm_variance} 
} &
\parbox[c][2.5cm][c]{6.5cm}{
CF\textsuperscript{}
} &
\parbox[c][2.5cm][c]{7.5cm}{
    \begin{equation}
  \begin{aligned}
    &\Sigma = \sum_{i=1}^{K}\pi_i(\Sigma_i+ {m_i \boldsymbol{\cdot} {m_i}^\top})\\
           &-{(\sum_{i=1}^{K}\pi_i m_i)\boldsymbol{\cdot} (\sum_{i=1}^{K}\pi_i m_i)^\top}
  \end{aligned}
	    \label{eq:variance_in_table}
\end{equation}
} \\
\hline
\parbox[c][2.5cm][c]{7.5cm}{
Empirical addition of \\physical quantities of the \\same type
} &
\parbox[c][2.5cm][c]{7.5cm}{
Addition of independent \acp{RV} \\corresponds with:\\[1em] Convolution of \acp{PDF}; see \\\Cref{prop:gmm_convolution} in \Cref{sec:appendix} 
} &
\parbox[c][2.5cm][c]{7.5cm}{
CF\textsuperscript{}
} &
\parbox[c][2.5cm][c]{7.5cm}{
    \begin{equation}
  \begin{aligned}
    &X+Y=Z, \quad X \perp Y\\
    &K_Z= K_X \boldsymbol{\cdot} K_Y, \quad {\pi_Z}_{ij} = {\pi_X}_i \boldsymbol{\cdot} {\pi_Y}_j\\
    &{m_Z}_{ij}= {m_X}_i + {m_Y}_j, \quad {\Sigma_Z}_{ij}= {\Sigma_X}_i + {\Sigma_Y}_j
  \end{aligned}
	    \label{eq:tab:convolution}
\end{equation}
} \\
\hline
\parbox[c][4.5cm][c]{6.5cm}{
Focusing on variables of \\interest
} &
\parbox[c][4.5cm][c]{6.5cm}{
Partial marginalization of \acp{RV}\\corresponds with:\\[1em]Integrating out nuisance \acp{RV};\\ see \Cref{prop:gmm_marginalization} in \Cref{sec:appendix}
} &
\parbox[c][4.5cm][c]{6.5cm}{
CF\textsuperscript{}
} &
\parbox[c][4.5cm][c]{7.2cm}{
    \begin{equation}
  \begin{aligned}
  &Z = \begin{bmatrix} X \\ Y \end{bmatrix} \in \mathbb{R}^d,\quad Z \sim \mathcal{G}_Z(z|g_Z)\\
  &{m_Z}_i = \begin{bmatrix} {{m_Z}_i}_X \\ {{{m_Z}}_i}_Y \end{bmatrix}, \quad
  {\Sigma_Z}_i = \begin{bmatrix}
{{\Sigma_Z}_{i}}_{XX}  {{\Sigma_Z}_{i}}_{XY} \\
{{\Sigma_Y}_{i}}_{YX}  {{\Sigma_Z}_{i}}_{YY}
\end{bmatrix}\\
&X    \sim   \mathcal{G}_X(x|g_X),\quad Y \sim   \mathcal{G}_Y(x|g_Y)\\
   &K_X = K_Y = K_Z\\
  &{\pi_X}_i = {\pi_Y}_i = {\pi_Z}_i\\
  &{m_X}_i  = {{m_Z}_i}_X,\quad {m_Y}_i  = {{m_Z}_i}_Y, \\
  &{\Sigma_X}_i  = {{\Sigma_Z}_i}_{XX},\quad {\Sigma_Y}_i  = {{\Sigma_Z}_i}_{YY}
  \end{aligned}
	    \label{eq:marginalization_in_table}
\end{equation}
} \\
\hline
\parbox[c][6.7cm][c]{6.5cm}{
Exploiting evidence \\about a set of \acp{RV}
} &
\parbox[c][6.7cm][c]{6.5cm}{
Partial conditioning of \acp{RV} \\regarding the observed \acp{RV};\\see \Cref{prop:gmm_conditioning} in \Cref{sec:appendix}
} &
\parbox[c][6.7cm][c]{6.5cm}{
CF\textsuperscript{}
} &
\parbox[c][6.7cm][c]{7.2cm}{
    \begin{equation}
  \begin{aligned}
&X|Y    \sim   \mathcal{G}_{X|Y}(x|y,g_{X|Y})\\   
&K_{X|Y}=K_Z\\
  &{m_{X|Y}}_i  = {m_i}_X+{\Sigma_{i}}_{XY}\boldsymbol{\cdot}{{\Sigma_{i}}_{YY}}^{-1}\boldsymbol{\cdot}(y-{m_i}_Y)\\   
  &{\Sigma_{X|Y}}_i  = {\Sigma_{i}}_{XX}-{\Sigma_{i}}_{XY}\boldsymbol{\cdot}{{\Sigma_{i}}_{YY}}^{-1}\boldsymbol{\cdot}{\Sigma_{i}}_{YX}\\
  &{\pi_{X|Y}}_i =\frac{\pi_i \mathcal{N}(y|{m_Y}_i,{\Sigma_Y}_i)}{\sum_{j=1}^{K_{Y}}\pi_j \mathcal{N}(y|{m_Y}_j,{\Sigma_Y}_j)}\\[1em]
  &Y|X    \sim   \mathcal{G}_{Y|X}(y|x,g_{Y|X})\\   
  & K_{Y|X}=K_Z\\
  &{m_{Y|X}}_i  = {m_i}_Y+{\Sigma_{i}}_{YX}\boldsymbol{\cdot}{{\Sigma_{i}}_{XX}}^{-1}\boldsymbol{\cdot}(x-{m_i}_x)\\   
  &{\Sigma_{Y|X}}_i  = {\Sigma_{i}}_{YY}-{\Sigma_{i}}_{YX}\boldsymbol{\cdot}{{\Sigma_{i}}_{XX}}^{-1}\boldsymbol{\cdot}{\Sigma_{i}}_{XY}\\
  &{\pi_{Y|X}}_i =\frac{\pi_i \mathcal{N}(x|{m_X}_i,{\Sigma_X}_i)}{\sum_{j=1}^{K_{X}}\pi_j \mathcal{N}(x|{m_X}_j,{\Sigma_X}_j)}
  \end{aligned}
	    \label{eq:conditioning_in_table}
\end{equation}
} \\
\hline
\parbox[c][4.5cm][c]{7.5cm}{
Mix of Sources
} &
\parbox[c][4.5cm][c]{7.5cm}{
Convex linear combination of \\\acp{PDF}; see \Cref{prop:gmm_convex_hull} in \\\Cref{sec:appendix}
} &
\parbox[c][4.5cm][c]{7.5cm}{
CF\textsuperscript{}
} &
\parbox[c][4.5cm][c]{7.5cm}{
    \begin{equation}
  \begin{aligned}
  &X^j \sim \mathcal{G}_{X^j}(x|g^j_X)\text{ for }j=1,\ldots,J\\
  &X \sim \mathcal{G}_X(x|g_X) =\sum_{j=1}^{J}w^j \mathcal{G}^{j}_{X}(x|g^j_X)=\\
  &\sum_{j=1}^{J}w^j \sum_{i=1}^{K_j} \pi^j_i\mathcal{N}_{X}(x|m^j_i,\Sigma^j_i)=\\ 
  &\sum_{j=1}^{J} \sum_{i=1}^{K_j} w^j \pi^j_i\mathcal{N}_{X}(x|m^j_i,\Sigma^j_i).
  \end{aligned}
	    \label{eq:source_mixture}
\end{equation}
} \\
\hline
\parbox[c][3.2cm][c]{6.5cm}{
Fusion of independent \\information contributions
} &
\parbox[c][3.2cm][c]{6.5cm}{
Bayesian fusion corresponds with:\\[1em] Multiplication of \acp{PDF}; see \\\Cref{prop:gmm_multiplication} in \Cref{sec:appendix}
} &
\parbox[c][3.2cm][c]{6.5cm}{
CF\textsuperscript{}
} &
\parbox[c][3.2cm][c]{7.2cm}{
{\tiny
    \begin{equation}
      \begin{aligned}
    &\mathcal{G}_X(x|g_a)\boldsymbol{\cdot}\mathcal{G}_X(x|g_b) = \sum_{i=1}^{K_a}\sum_{j=1}^{K_b}c_{a_i b_j}{\pi_a}_i{\pi_b}_j \mathcal{N}_X(x|{m_{a_ib_j}},{\Sigma_{a_ib_j}})\\
    &c_{a_ib_j} = \frac{\exp\left[-\frac{1}{2}(m_{a_i}-m_{b_j})^\top \boldsymbol{\cdot}(\Sigma_{a_i}+\Sigma_{b_j})^{-1}\boldsymbol{\cdot}(m_{a_i}-m_{b_j})\right] }{\sqrt{\det{(2\pi(\Sigma_{a_i}+\Sigma_{b_j}))}}}\\
    &m_{a_ib_j}=  (\Sigma_{a_i}^{-1}+\Sigma_{b_j}^{-1})^{-1} \boldsymbol{\cdot}(\Sigma_{a_i}^{-1}\boldsymbol{\cdot}m_{a_i}+\Sigma_{b_j}^{-1}\boldsymbol{\cdot}m_{b_j})\\
      \end{aligned}
	    \label{eq:bayesian_fusion}
    \end{equation}\\[-4.5em]
    \begin{equation}
    \Sigma_{a_ib_j}= (\Sigma_{a_i}^{-1}+\Sigma_{b_j}^{-1})^{-1}
    \end{equation}
}
} \\
\hline
\parbox[c][2.0cm][c]{6.5cm}{
Seeing the outcome of a \\\ac{GMM} distributed \ac{RV}
} &
\parbox[c][2.0cm][c]{6.5cm}{
Drawing samples from a \\\ac{GMM} distributed \ac{RV}
} &
\parbox[c][2.0cm][c]{6.5cm}{
N\textsuperscript{}
} &
\parbox[c][2.0cm][c]{7.2cm}{
Randomly select a component using the mixture weights and sample from its Gaussian distribution. 
See \Cref{prop:multivariate_gaussian_sampling} and \Cref{prop:gmm_sampling} in \\\Cref{sec:appendix}.
} \\
\hline
\parbox[c][7.4cm][c]{7.3cm}{
Reducing the number\\ of components
} &
\parbox[c][7.4cm][c]{7.3cm}{
Reducing the number of \\parameters of \acp{GMM} based on \\$\Vert \boldsymbol{\cdot} \Vert_{L^2(\mathbb{R}^d)}.$
} &
\parbox[c][7.4cm][c]{7.3cm}{
N\textsuperscript{}
} &
\parbox[c][7.4cm][c]{7.3cm}{
{\tiny
            \begin{equation}
              \begin{split}
            &\argmin_{g_b}^{} \Vert \mathcal{G}_X(x|g_a)- \mathcal{G}_X(x|g_b) \Vert_{L^2(\mathbb{R}^d)} \text{ with } K_a >K_b.\\
            &\argmin_{g_b}^{} \Vert \mathcal{G}_X(x|g_a)- \mathcal{G}_X(x|g_b) \Vert_{L^2(\mathbb{R}^d)} =\\
            &\argmin_{g_b}^{} \sqrt{{\int_{\mathbb{R}^d}^{} {\left(\mathcal{G}_X(x|g_a)-\mathcal{G}_X(x|g_b)\right)}^2\mathrm{d}x}}=\\
            &\argmin_{g_b}^{} \sqrt{I_{aa}-2I_{ab}+I_{bb}}\\
            &I_{aa}=\sum_{i=1}^{K_a}\sum_{j=1}^{K_a}{\pi_{a_i}}{\pi_{a_j}} c_{a_i a_j}\\
            &I_{ab}=\sum_{i=1}^{K_a}\sum_{j=1}^{K_b}{\pi_{a_i}}{\pi_{b_j}} c_{a_i b_j}\\
            &I_{bb}=\sum_{i=1}^{K_b}\sum_{j=1}^{K_b}{\pi_{b_i}}{\pi_{b_j}} c_{b_i b_j}\\
            &c_{uv}= \frac{\exp\left[-\frac{1}{2}(m_u-m_v)^\top \boldsymbol{\cdot}(\Sigma_u+\Sigma_v)^{-1}\boldsymbol{\cdot}(m_u-m_v)\right] }{\sqrt{\det{(2\pi(\Sigma_u+\Sigma_v))}}}\\
            &m_{uv}=  (\Sigma_u^{-1}+\Sigma_v^{-1})^{-1} \boldsymbol{\cdot}(\Sigma_u^{-1}\boldsymbol{\cdot}m_u+\Sigma_v^{-1}\boldsymbol{\cdot}m_v)\\
            &\Sigma_{uv}= (\Sigma_u^{-1}+\Sigma_v^{-1})^{-1}\\
	          &u,v \in \{a_i\}_{i=1,\ldots,K_a} \cup \{b_j\}_{j=1,\ldots,K_b}
              \end{split}
              \label{eq:table_l2_distance_gmm}
        \end{equation}
}
} \\
\hline
\end{longtable}
\clearpage

}

\section{Examples}
\label{sec:examples}
Now we present some examples to demonstrate the practical usefulness of representing quantitative attributes by the means of \acp{GMM} 
and to calculate with them according to the formulas of \Cref{tab:gm_math}.

\subsection{Mechanical Assembly}
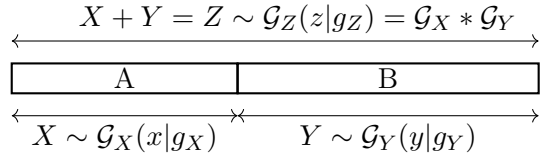
\begin{figure}[H]
    \centering
    \resizebox{0.45\textwidth}{!}{\begin{tikzpicture}
  \draw[thick] (0,0) rectangle (3,0.4);
  \draw[thick] (3,0) rectangle (7,0.4);

  \draw[<->] (0,-0.3) -- (3,-0.3);
  \node at (1.5,-0.6) {$X \sim \mathcal{G}_X(x|g_X)$};
  \draw[<->] (3,-0.3) -- (7,-0.3);
  \node at (5.0,-0.6) {$Y \sim \mathcal{G}_Y(y|g_Y)$};

  \draw[<->] (0,0.7) -- (7,0.7);
  \node at (3.8,1.0) {$X+Y=Z \sim \mathcal{G}_Z(z|g_Z) = \mathcal{G}_X * \mathcal{G}_Y $};

  \node at (1.5,0.2) {A};
  \node at (5.0,0.2) {B};

\end{tikzpicture}}
    \caption{Length $X$ and length $Y$ make up the total length $Z$. $\mathcal{G}_Z$ can be calculated using \Cref{eq:tab:convolution} in \Cref{tab:gm_math}. The symbol $*$ denotes the convolution operator.}
    \label{fig:additive_length}
\end{figure}

In assembly situations parts are put together in a way that it results in an
additive behavior. The lengths are modeled as \acp{RV}, when the part $A$ with length $X \sim \mathcal{G}_X(x|g_X)$ and
the part $B$ with length $Y \sim \mathcal{G}_Y(y|g_Y)$ are aligned according
to \Cref{fig:additive_length} to make up the total length $Z$. In general for stochastically independent \acp{RV} $X \sim p_X(x)$ and $Y \sim p_Y(y)$, 
the \ac{PDF} of the total length $Z$ is given by the convolution of the \acp{PDF} of $X$ and $Y$:
\begin{equation}
Z = X + Y \text{ and }X \perp Y \implies p_Z(z) =\int_{\Omega}^{}p_X(x)\, p_Y(z - x)\, \mathrm{d}x= p_X(x) * p_Y(y).
\end{equation}

However, when the \acp{PDF} of the \acp{RV} are distributed according to \acp{GMM} the total length can be
calculated according to \Cref{eq:tab:convolution} in \Cref{tab:gm_math} in an arithmetical fashion without any approximation on the \ac{PDF} of the \ac{RV} $Z$.

\begin{figure}[H]
    \centering
    \includegraphics[width=1.0\textwidth]{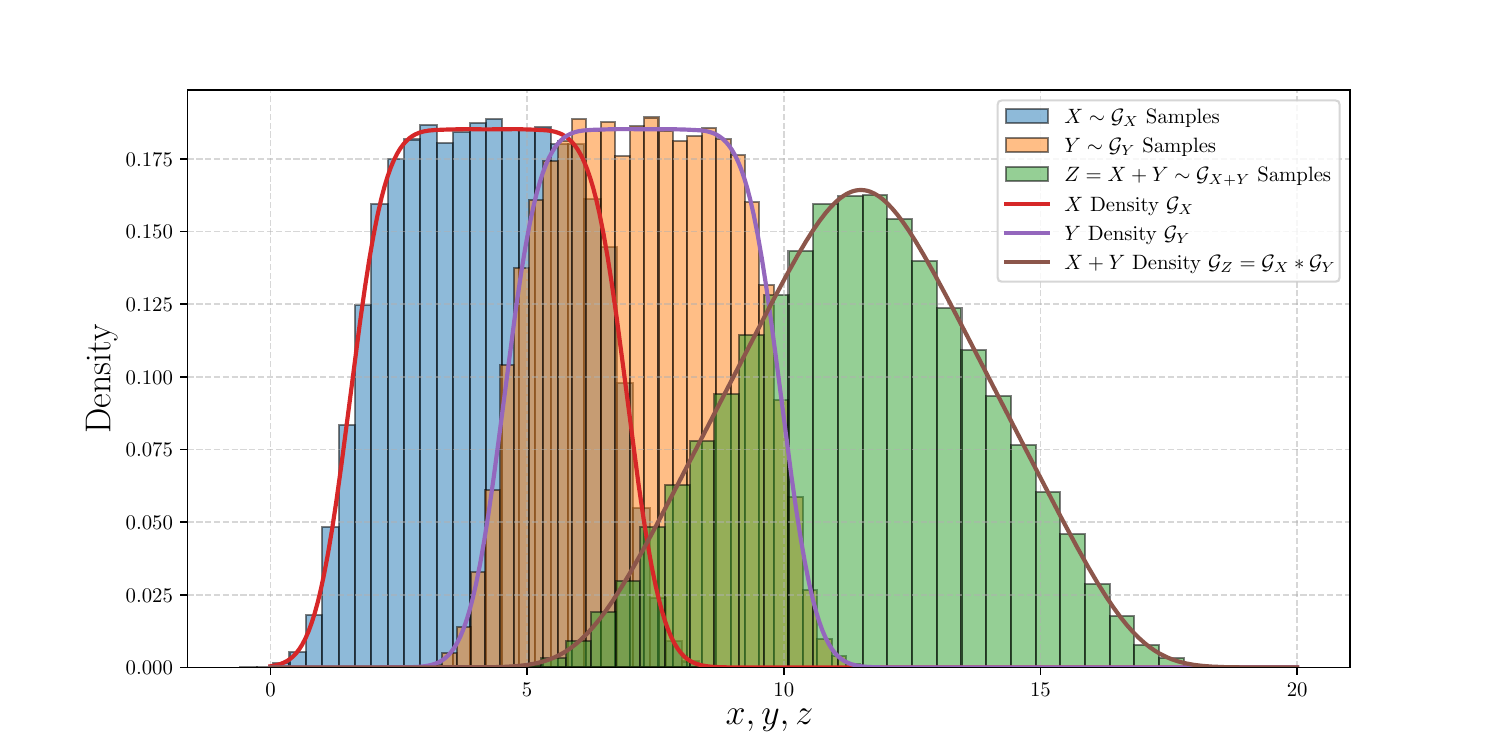}
    \caption{Analytical calculation of the resulting \ac{PDF} vs forward Monte Carlo according to \cite{JCGM2008} using the ad hoc sampler in \Cref{sec:gmm_sampling}}
    \label{fig:additive_length_sampling_vs_analytical}
\end{figure}

As a numerical example consider two six components \acp{GMM} for the two \acp{RV} $X$ and $Y$ according
to \Cref{fig:additive_length}. It can be seen in \Cref{fig:additive_length_sampling_vs_analytical} that the closed form arithmetic form \ac{PDF} assigned to the \ac{RV} $Z$
coincides with the histogram resulted from samples achieved using the forward Monte Carlo calculation \cite{JCGM2008}.

\begin{remark}
	It is possible to apply \Cref{eq:tab:convolution} successively to treat the countable summation
    cases for example \Cref{fig:resistances_series} similarly.
    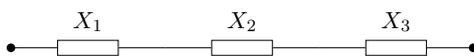
\begin{figure}[H]
    \centering
    \resizebox{0.4\textwidth}{!}{\begin{tikzpicture}[circuit ee IEC]
    \fill (0.0,0.0) circle (2pt);
    \fill (7.5,0.0) circle (2pt);
    \draw (0.0,0.0) to [resistor={info=$X_1$}] (2.5,0.0);
    \draw (2.5,0.0) to [resistor={info=$X_2$}] (5.0,0.0);
    \draw (5.0,0.0) to [resistor={info=$X_3$}] (7.5,0.0);
\end{tikzpicture}}
    \caption{Three independent resistances $X_1,X_2$ and $X_3$ make up the total resistance $Z=X_1 + X_2 + X_3$. Then for $X_i \sim \mathcal{G}_{X_i}, i=1,2,3$ the total resistance $Z \sim \mathcal{G}_Z = \mathcal{G}_{X_1} * \mathcal{G}_{X_2} * \mathcal{G}_{X_3}$ which can be calculated by applying \Cref{eq:tab:convolution} successively.}
    \label{fig:resistances_series}
\end{figure}
\end{remark}

\subsection{Mechanical Disassembly}
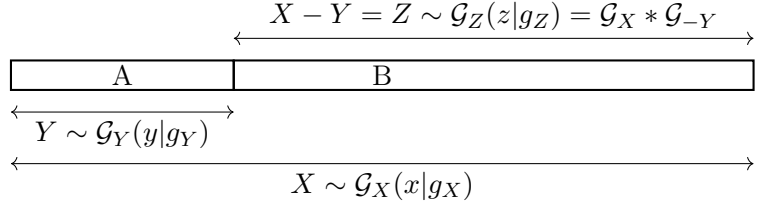
\begin{figure}[H]
    \centering
    \resizebox{0.63\textwidth}{!}{\begin{tikzpicture}
  \draw[thick] (0,0) rectangle (3,0.4);
  \draw[thick] (3,0) rectangle (10,0.4);

  \draw[<->] (0,-0.3) -- (3,-0.3);
  \node at (1.5,-0.6) {$Y \sim \mathcal{G}_Y(y|g_Y)$};
  \draw[<->] (3,+.7) -- (10,+.7);
  \node at (6.5,+1.0) {$X-Y=Z \sim \mathcal{G}_Z(z|g_Z) = \mathcal{G}_X * \mathcal{G}_{-Y} $};

  \draw[<->] (0,-1.0) -- (10,-1.0);
  \node at (5.0,-1.3) {$X \sim \mathcal{G}_X(x|g_X)$};

  \node at (1.5,0.2) {A};
  \node at (5.0,0.2) {B};

\end{tikzpicture}}
    \caption{Length $Y$ and length $Z$ make up the total length $X$. $\mathcal{G}_Z$ can be calculated using \Cref{eq:tab:convolution} in \Cref{tab:gm_math}. The symbol $*$ denotes the convolution operator.}
    \label{fig:subtractive_length}
\end{figure}

In disassembly situations parts are taken apart in a way that it would result in a
subtractive behavior. The lengths are modeled as \acp{RV}, when the part $A$ with length $Y \sim \mathcal{G}_Y(y|g_Y)$ and
the part $B$ with length $Z \sim \mathcal{G}_Z(z|g_Z)$ are aligned according
to \Cref{fig:subtractive_length} to make up the total length $X$. In general for stochastically independent \acp{RV} $X \sim p_X(x)$ and $Y \sim p_Y(y)$, 
the \ac{PDF} of the subtracted length $Z$ is given by the convolution of the \acp{PDF} of $X$ and $-Y$. In the case of \acp{GMM} when $Y \sim \mathcal{G}_Y(y|g_Y)$ then $-Y \sim \mathcal{G}_{-Y}(y|g_{-Y})$ where $g_{-Y}$ is the same as $g_Y$ except that the means are negated.

\subsection{Metal Sheet}
\label{sec:metal_sheet}
\begin{figure}[H]
    \centering
    \resizebox{0.5\textwidth}{!}{\begin{tikzpicture}
  \draw (0,0) rectangle (4,2);

  \draw[<->] (0,-0.3) -- (4,-0.3);
  \node at (2,-0.6) {$X \sim \mathcal{G}_X(x|g_X)$};

  \draw[<->] (-0.3,0) -- (-0.3,2);
  \node at (-1.7,1) {$Y \sim \mathcal{G}_Y(x|g_Y)$};

  \node at (2,1) {$Z = X \boldsymbol{\cdot} Y$};
\end{tikzpicture}}
    \caption{A rectangle with side lengths $X \sim \mathcal{G}_X(x|g_X)$, $Y \sim \mathcal{G}_Y(x|g_Y)$, and area $Z=X \boldsymbol{\cdot} Y$.}
    \label{fig:rectangle}
\end{figure}
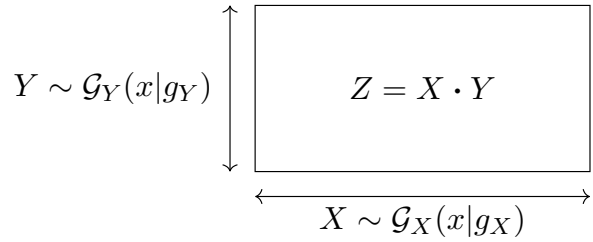

The mass of a rectangle sheet of metal in \Cref{fig:rectangle} is proportional to the area of the rectangle having side lengths
$X \sim \mathcal{G}_X(x|g_X)$ and $Y \sim \mathcal{G}_Y(x|g_Y)$. The area is the result of a multiplication $Z=X\boldsymbol{\cdot}Y$.

In this case the \ac{PDF} of $Z$ cannot be calculated in closed form even if
$X$ and $Y$ are distributed according to simple \acp{PDF}~\cite{DasGupta2010}. 
In general if $X \sim p(x)$ and $Y \sim p(y)$ and $Z \sim p(z)$:
\begin{equation}
	Z = X \boldsymbol{\cdot} Y \text{ and } X \perp Y \implies p_Z(z) = \int_{\Omega}^{} \frac{1}{|x|} p_X(x)p_Y\left(\frac{z}{x}\right)\,\mathrm{d}x.
\end{equation}
As it can be deduced from the equation, the parameters of $Z \sim \mathcal{G}_Z(z|g_Z)$ cannot be calculated using an arithmetical procedure. The \ac{PDF} of $Z$ is a
complicated distribution in terms of the \acp{PDF} of $X$ and $Y$; called ``the product distribution'' \cite{DasGupta2010}. However, 
it is still possible to fit a \ac{GMM} to $Z$ using the samples achieved from the forward
Monte Carlo procedure~\cite{JCGM2008}. For the sampler one can use the ad hoc sampler discussed in \Cref{sec:gmm_sampling} or a general sampler like~\cite{Hoffman2011}.

\begin{figure}
    \centering
    \includegraphics[width=1.0\textwidth]{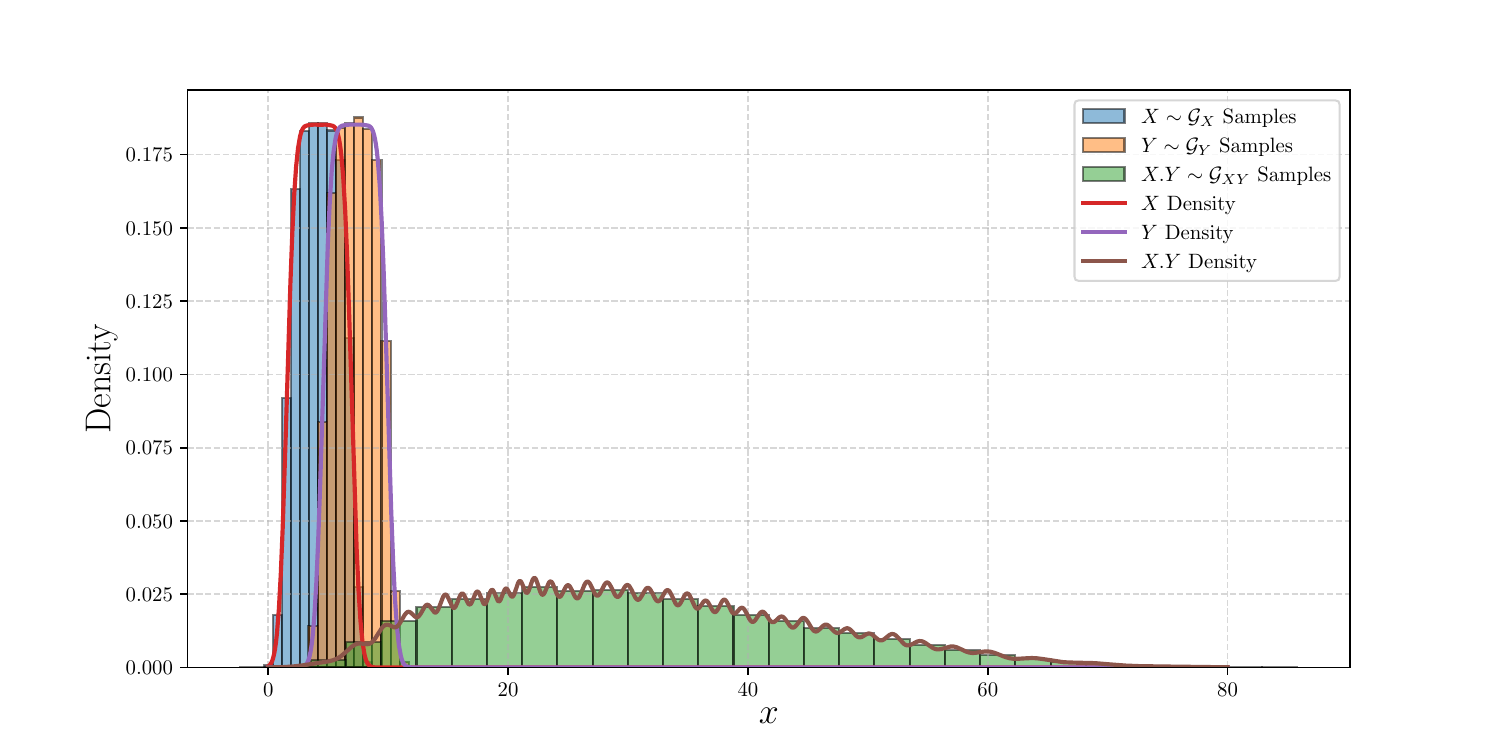}
    \caption{Numerical calculation of the resulting \ac{PDF} using the \ac{EM} algorithm based on the samples produced using the forward Monte Carlo according to \cite{JCGM2008} using the ad hoc sampler in \Cref{sec:gmm_sampling}}
    \label{fig:multiplicative_area}
\end{figure}

As a numerical example consider two six components \acp{GMM} for the two \acp{RV} $X$ and $Y$ according to
 \Cref{fig:multiplicative_area}. The \ac{EM} algorithm needs the number of components that is to be fit to be fixed.
Here a plausible heuristic would be $K=36$ as in the special case where all the standard deviations were zero the resulting product would only assume at maximum $36$ different distinct values. It can be seen
 in \Cref{fig:multiplicative_area} that the \ac{PDF} assigned to the \ac{RV} $Z$ coincides with the histogram of the samples achieved using the forward Monte Carlo \cite{JCGM2008}.

\subsection{Mix of Sources}
In many production and manufacturing processes outsourcing must be done. Assume parts are to be delivered by two suppliers
$A$ and $B$. Assume the part is a shaft and the value of interest to the main manufacturer is the diameter of the shaft $X$ and he
is interested to model the ensemble of shafts they are receiving by a \ac{RV} and assign a \ac{PDF} in the form of a \ac{GMM} to it.
Due to different machineries of the suppliers one cannot assume symmetry and must come up with a systematic way of assigning 
the \ac{GMM}. If supplier $A$ delivers $r_A\in[0,100]\%$ of the shafts and supplier $B$ supplies $r_B\in[0,100]\%$ of the shafts and
the diameter of the shafts supplied by $A$ is to be modelled by the \ac{RV} $X \sim \mathcal{G}_{X}(x|g_{X_A})$ and the diameter of the shafts supplied by $B$ is to be modelled by the \ac{RV} $X \sim \mathcal{G}_{X}(x|g_{X_B})$ then the diameter of the shafts supplied to the main manufacturer can be modelled by the \ac{RV} $X \sim \mathcal{G}_{X}(x|g_{X_S})$ where:
\begin{equation}
    \mathcal{G}_{X}(x|g_{X_S}) = \frac{r_A}{r_A+r_B} \mathcal{G}_X(x|g_{X_A}) + \frac{r_B}{r_A+r_B} \mathcal{G}_{X}(x|g_{X_B}).
\end{equation}

The resulting distribution is again a Gaussian mixture model with $K_S = K_A + K_B$ number of components. 
Using a single combined index $k$, the resulting mixture can be written as
\begin{equation}
\mathcal{G}_{X_S}(x|g_{X_S}) = \sum_{k=1}^{K_S} {\pi_{S}}_k\mathcal{N}(x|{m_{S}}_k, {\Sigma_{S}}_k) .
\end{equation}

The parameters of the resulting mixture are obtained as follows:
\begin{equation}
	\begin{split}
		{\pi_{S}}_k &= \frac{r_A}{r_A+r_B} {\pi_{A}}_k,\quad{m_{S}}_k = {m_{A}}_k,\quad{\Sigma_{S}}_k = {\Sigma_{A}}_k \text{ for } k = 1,\ldots,K_A\\
		{\pi_{S}}_k &= \frac{r_B}{r_A+r_B} {\pi_{B}}_{k-K_A},\quad{m_{S}}_k = {m_{B}}_{k-K_A},\quad{\Sigma_{S}}_k = {\Sigma_{B}}_{k-K_A} \text{ for } k = K_A+1,\ldots,K_S.
	\end{split}
\end{equation}

Thus, an implementation only requires concatenating the Gaussian components of both mixtures and rescaling the mixture weights according to the supplier proportions.

This is consistent both with the sampling procedure of Monte Carlo methods \cite{Robert2004} and also the functional analysis point of view of the probability theory \cite{Klenke2020}.
The beauty of it is that the mixture of two \acp{GMM} is again a \ac{GMM} and the parameters of the resulting \ac{GMM} can be calculated in a arithmetical procedure and closed form; see \Cref{prop:gmm_convex_hull} in \Cref{sec:appendix}.

\begin{remark}
Circular factories use partly old parts and products degraded 
varyingly from different generations and different durations of
 use. Therefore, the mixing of sources is an inevitable operation in circular 
factories and must be handled properly.
\end{remark}

\subsection{Probabilistic Quality Control}
\label{sec:quality_control}
\begin{figure}[H]
    \centering
    \resizebox{0.25\textwidth}{!}{\begin{tikzpicture}[scale=1.0]

\draw[->] (0,0) -- (2,0) node[right] {$x_1$};
\draw[->] (0,0) -- (0,2) node[above] {$x_2$};

\draw[pattern=north west lines] (0.2,1.2) rectangle (1.2,1.7);
\node at (0.7,1.9) {$Q_r$};
\draw[ rotate=+23,pattern=north west lines] (1.9,0.1) ellipse (0.3 and 0.5);
\node at (2.2,1.0) {$Q$};

\end{tikzpicture}}
    \caption{If quality is defined by the set $Q$, the probability of the attributes 
	of the product instance to fall into set $Q$ can be calculated by integrating the \ac{PDF} 
	over the set $Q$. Tolerances are usually given as intervals for the components $x_i$ of 
	the product instance attribute vector $x$, are a special case of $Q$, where $Q$ 
	becomes the hyperrectangle $Q_r$. If the tolerances of the components of the 
	product instance attribute vector $x$, are correlated (not independent), $Q$ 
	can assume a general shape.}
    \label{fig:probabilistic_quality_control}
\end{figure}
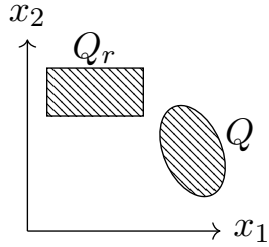

Let the vector of the product attributes be denoted by $x \in \mathbb{R}^d$ which is the realization of the \ac{RV} $X \sim \mathcal{G}_X(x|g)$.
The quality engineer specifies an acceptable region in form of a set $Q \subset \mathbb{R}^d$ according to 
the \Cref{fig:probabilistic_quality_control} in the attribute space, representing 
the set that the features are allowed to fall in. The probability that the attribute vector of the product to fall into the allowed set $Q$ is:

\begin{equation}
\mathbb{P}(x \in Q) = \int_{Q} \mathcal{G}_X(x|g)\mathrm{d}x .
\end{equation}

As sampling from a \ac{GMM} is straightforward; Monte Carlo methods \cite{Robert2004} 
can be used to estimate the probability of the attribute vector of the product instance to fall into the set $Q$.

\subsection{Dependent Variables}
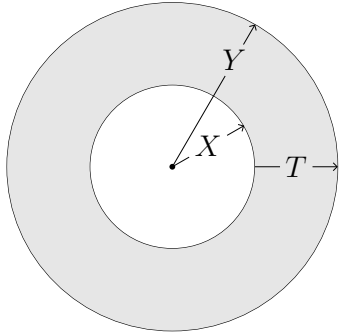
\begin{figure}[!ht]
    \centering
    \resizebox{0.35\textwidth}{!}{    \begin{tikzpicture}[> = {Straight Barb[scale=1.2]}
                        ]
\draw[draw, double = gray!20, double distance=2cm] (0,0) circle [radius=3];
%
\path (0,0) coordinate (o) to node (lbl) {\huge $X$} ++(30:2);
\draw[->] (o) -- (lbl) -- (30:2);
\path (o) to node[pos=0.75] (lbl) {\huge $Y$} ++(60:4);
\draw[->] (o) -- (lbl) -- (60:4);
\path ++(0:2) to node[pos=0.25] (lbl) {\huge $T$} ++(0:4);
\draw[->] (0:2) -- (lbl) -- (0:4);

\fill   (0,0) circle [radius=2pt] node [above left] {$$};
\end{tikzpicture}}
    \caption{A pipe with the inner diameter $X$ and the outer diameter $Y$. The thickness of the pipe is $T$. The variables are modeled as \acp{RV}.}
    \label{fig:pipe}
\end{figure}

Assume the inner and outer radii of the pipe in \Cref{fig:pipe} are modeled as 
stochastically dependent \acp{RV} and their 
relation to each other is modeled by the \ac{RV}
$Z = \left[\begin{smallmatrix} X \\ Y \end{smallmatrix}\right] \in \mathbb{R}^2$, $Z \sim \mathcal{G}_Z(z|g_Z)$. Two common 
questions could arise:
\begin{enumerate}
    \item One might be interested in having an estimate of the inner radius $X$ without observing, measuring, or looking at the outer 
    radius. Then $Y$ is a nuisance variable which is integrated out (marginalization) using \Cref{prop:gmm_marginalization} in \Cref{sec:appendix}. For 
	\acp{GMM} this means that the nuisance variable is simply dropped in the joint \ac{GMM} of $X$ and $Y$, i.e. in $\mathcal{G}_Z$ 
	to obtain the \ac{PDF} of $X$.
	\item One might be interested in having an estimate of the inner radius 
$X$ after observing the outer radius, mathematically speaking putting $Y=y^*$ (conditioning), which can be 
handled using \Cref{prop:gmm_conditioning} in \Cref{sec:appendix}.
\end{enumerate}

Both cases can be treated in closed form without any approximation and in an arithmetical fashion.

\subsection{Fusion of Information from Multiple Measurement Devices}
Let $X \sim \mathcal{G}_X(x|g_A)$ and $X \sim \mathcal{G}_X(x|g_B)$ be 
the measurement results of a physical quantity $x$ 
of interest done by two stochastically independent apparatuses $A$ and $B$ respectively. To fuse these 
information one must multiply the two \acp{PDF} according to the Bayes' theorem, and normalize 
the result to arrive at a valid \ac{PDF}. \Cref{eq:gmm_multiplication} gives a closed form formula 
to calculate the normalization constant. The derivation 
does not comprise any approximations and is done in an arithmetical fashion.

\subsection{Reduction of Gaussian Mixture Models}
In \acp{GMM}, the number of parameters increases rapidly under operations such as 
Bayesian fusion~(multiplying two mixtures) or convolution, since each pairwise 
interaction between components typically produces new mixture components. 
Multiplying or convolving two \acp{GMM} with $K$ and $L$ components result in a \ac{GMM} 
with $M=K\boldsymbol{\cdot}L$ components. With respect to the number of such operations, the number 
of components of \ac{GMM} grows exponentially.
Therefore, it is essential to perform mixture reduction to maintain tractability. 
A key advantage of working with Gaussian mixtures is that the $L^2$ distance 
between mixtures admits a closed-form expression~\cite{Huber2015}, because products and 
integrals of Gaussian functions can be calculated analytically. As a result, 
instead of solving an infinite dimensional functional optimization problem over 
arbitrary densities, we can reformulate mixture reduction as a finite dimensional 
optimization problem over the mixture weights, means, and covariances. This 
significantly improves computational feasibility while preserving a principled approximation framework. 
The closed form formula for the $L^2$ distance between two Gaussian mixtures is:

\begin{equation}
	\label{eq:l2_distance_gmm}
	{\Vert \mathcal{G}_X(x|g_a)-\mathcal{G}_X(x|g_b) \Vert}_{{L^2}{(\mathbb{R}^d})}= \sqrt{{\int_{\mathbb{R}^d}^{} {\left(\mathcal{G}_X(x|g_a)-\mathcal{G}_X(x|g_b)\right)}^2\mathrm{d}x}}
	= \sqrt{I_{aa}-2I_{ab}+I_{bb}}
\end{equation}
where
\begin{equation}
	\begin{split}
		I_{aa}&=\sum_{i=1}^{K_a}\sum_{j=1}^{K_a}{\pi_{a_i}}{\pi_{a_j}} c_{a_i a_j}\\
		I_{ab}&=\sum_{i=1}^{K_a}\sum_{j=1}^{K_b}{\pi_{a_i}}{\pi_{b_j}} c_{a_i b_j}\\
		I_{bb}&=\sum_{i=1}^{K_b}\sum_{j=1}^{K_b}{\pi_{b_i}}{\pi_{b_j}} c_{b_i b_j}
    \end{split}
\end{equation}
and
\begin{equation}
	\begin{split}
		c_{uv}&= \frac{\exp\left[-\frac{1}{2}(m_u-m_v)^\top \boldsymbol{\cdot}(\Sigma_u+\Sigma_v)^{-1}\boldsymbol{\cdot}(m_u-m_v)\right] }{\sqrt{\det{(2\pi(\Sigma_u+\Sigma_v))}}}\\
    	m_{uv}&=  (\Sigma_u^{-1}+\Sigma_v^{-1})^{-1} \boldsymbol{\cdot}(\Sigma_u^{-1}\boldsymbol{\cdot}m_u+\Sigma_v^{-1}\boldsymbol{\cdot}m_v)\\
    	\Sigma_{uv}&= (\Sigma_u^{-1}+\Sigma_v^{-1})^{-1}\\
		u,v &\in \{a_i\}_{i=1,\ldots,K_a} \cup \{b_j\}_{j=1,\ldots,K_b}.
  	\end{split}
\end{equation}

For a given \ac{GMM} $\mathcal{G}_X(x|g_a)$ with $K_a$ components the reduction to a \ac{GMM} 
with $K_b < K_a$ components can be accomplished by solving the optimization problem:
\begin{equation}
	\argmin_{g_b}^{} \Vert \mathcal{G}_X(x|g_a)- \mathcal{G}_X(x|g_b) \Vert_{L^2(\mathbb{R}^d)} \text{ with } K_a > K_b.
\end{equation}

\begin{figure}[H]
    \centering
    \resizebox{0.75\textwidth}{!}{\begin{tikzpicture}[scale=1.0]

\draw[->] (-0.5,-2.5) -- (3.5,-2.5) node[right] {$x_1$};
\draw[->] (-0.5,-2.5) -- (-0.5,1.5) node[above] {$x_2$};
\draw[->] (7+-0.5,-2.5) -- (7+3.5,-2.5) node[right] {$x_1$};
\draw[->] (7+-0.5,-2.5) -- (7+-0.5,1.5) node[above] {$x_2$};


\draw[ rotate=+0,pattern=north west lines] (0.0,0.0) ellipse (0.3 and 1.0);
\draw[ rotate=+0,pattern=north east lines] (0.0,-1.0) ellipse (0.3 and 1.0);
\draw[ rotate=+0,pattern=north west lines] (1.0,-2.0) ellipse (1.0 and 0.3);
\draw[ rotate=+0,pattern=north east lines] (2.0,-2.0) ellipse (1.0 and 0.3);

\draw[rotate=+0] (7+1.5,-2.0) ellipse (1.6 and 0.4);
\draw[rotate=+0] (7+0.0,-0.5) ellipse (0.4 and 1.6);

\end{tikzpicture}}
    \caption{Reduction of a \ac{GMM} with four components (of the left) into a \ac{GMM} with two components (on the right); illustration.}
    \label{fig:reduction}
\end{figure}
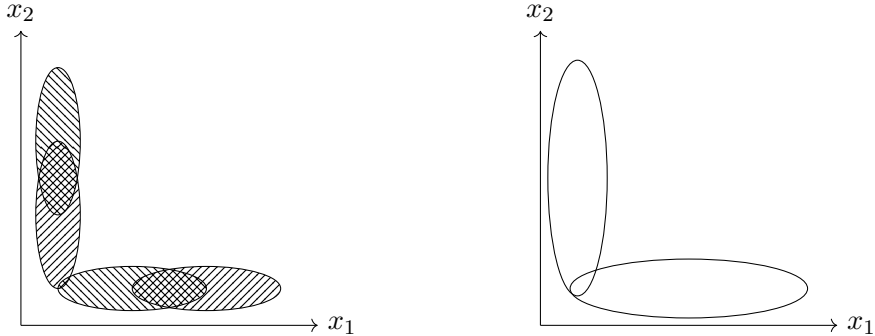

\Cref{fig:reduction} shows an illustration on how a \ac{GMM} having four components 
can be approximated with a \ac{GMM} having two components. According 
to \Cref{sec:small_footprint} and since $d=2$ now we have $11$ parameters instead of $23$ parameters after the reduction.

An overview of common reduction methods for \acp{GMM} can be found in~\cite{Fischer2026}\cite{Huber2015}.

\section{Modeling of Measurement Systems with \acp{GMM}}
\label{sec:measurement_system}

Abstractly, a measuring instrument can be described probabilistically by a joint \ac{PDF} $p(x,y)$ over the quantities to
be measured $x$ (measurands) and the observable data $y$ (observables).
$p(y|x) = \frac{p(x,y)}{p(x)}$ is the so-called forward model, which describes the
\ac{PDF} of the observables $y$ depending on the underlying value of
the measurand. If concrete data $y^*$ are observed and inserted into $p(y|x)$,
this becomes the likelihood function $p(y^*|x)$, which is now interpreted as a
function of $x$. According to Bayes' formula, the posterior \ac{PDF} of $x$
can be calculated in the form of  $p(x|y^*) = \frac{p(y^*|x)p(x)}{p(y^*)}$~\cite{Beyerer1999}.
Measuring has the goal to infer the underlying value of $x$ based on concretely
observed data $y^*$. The question is: What should be reported from the data $y^*$ in
the best possible way with respect to $x$? If one wants to keep the loss of
information to a minimum in the sense of Shannon's information theory, it is
optimal to calculate the posterior \ac{PDF} $p(x|y^*)$ and report it as a result
of the processing of $y^*$~\cite{Kotelnikov1968}\cite{Winkler1977}. If $p(x,y)$ is
modeled in the form of a \ac{GMM}, which is possible with arbitrary precision due
to the universal approximation property of \acp{GMM}~\cite{Mazja2014}\cite{Lu2017},
then $p(x|y^*) \propto p(x,y^*)$ is itself a \ac{GMM}, whose parameters can be
algebraically calculated from the parameters of the $p(x,y)$ \ac{GMM}. If one
also assumes an unbiased measuring device for which one can assume the prior
\acp{PDF} $p(x)$ is approximately uniformly distributed over the $\mathrm{Range}(x)$. 

\begin{figure}
     \centering
     \includegraphics[width=1.0\textwidth]{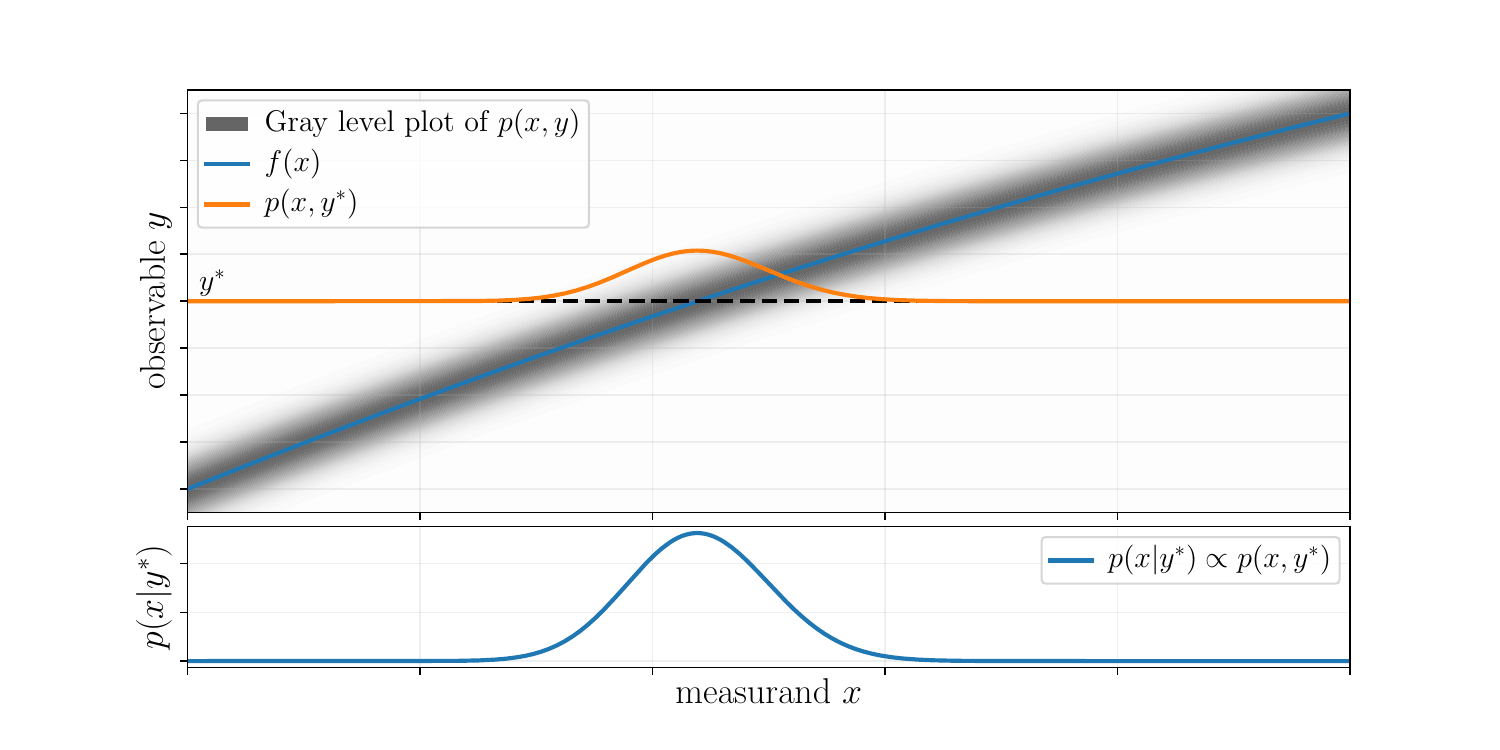}
     \caption{Schematic illustration of the conditioning process to obtain the posterior \ac{PDF}~$p(x|y^*)$ of the measurand based on an observation of the measurement device.}
     \label{fig:conditional_schematic}
 \end{figure}

Note that for measurement systems that have, up to noise, an approximately linear characteristic between $y$ and $x$, the marginal \ac{PDF} $p(y)$ is approximately uniformly distributed as well. 
Then $p(x|y^*) \propto p(y^*|x)$, i.e. proportional to
the likelihood function $p(y^*|x)$. For illustration see \Cref{fig:conditional_schematic}. 

Formulas for conditioning of \acp{GMM} can be found in \Cref{prop:gmm_conditioning} in \Cref{sec:appendix}.
Thus, not only continuous attributes itself can be
comprehensively described by means of \ac{GMM} in the probabilistic
sense, but also measurement systems in general as well as the processing
of observed (read off) data $y*$ in order to calculate the measurement
result in the form of the posterior \ac{PDF} $p(x|y*)$. Interestingly,
all required descriptions and calculations can be accomplished
within the function set of \acp{GMM}. The following example is intended to illustrate the interrelations.

\subsection{Model Selection}

When fitting a \ac{GMM}, the number of mixture components $K$ is typically unknown and 
must be estimated from the data. Two widely used criteria for this purpose are the 
\ac{AIC} and the \ac{BIC}~\cite{Bishop2019}. Both criteria balance goodness of fit and model complexity. 
For a candidate model with $K$ components, we first fit the \ac{GMM} and compute the 
maximized log-likelihood $\log L$. The criteria are then calculated as
\begin{equation}
\text{AIC} = -2 \log L + 2p, \qquad
\text{BIC} = -2 \log L + p \log N, \qquad
L  = \prod_{n=1}^{N} \mathcal{G}_X(x_n|g_X)
\end{equation}
where $p$ is the number of free parameters in the mixture model and $N$ is 
the number of observations. Increasing $K$ typically improves the likelihood but 
also increases $p$, which both criteria penalize. By fitting models for several 
values of $K$ and selecting the one that minimizes \ac{AIC} or \ac{BIC}, we obtain 
a principled estimate of the number of mixture components. In practice, 
\ac{BIC} often favors simpler models because its penalty grows with 
sample size $N$, whereas \ac{AIC} may select slightly larger mixtures.

\subsection{Example}
Assume the simplest case where a scalar input $x$ is to be measured by a 
measurement device and the output is a scalar observable $y$ which is 
corrupted by additive Gaussian noise. The physical laws governing the 
measurement device will manifest as the measurement curve $f(x)$. In short:

\begin{equation}
\begin{split}
&f \colon \mathbb{R} \to \mathbb{R} \qquad x \mapsto f(x)\\
&\widetilde{f} \colon \mathbb{R}^2 \to \mathbb{R} \quad x,w \mapsto y= f(x) + w\\
&W \sim \mathcal{N}(w|0,\sigma^2). \\
\end{split}
\end{equation}

Now, we assume that we have observed $N$ pairs of $\{(x_i,y_i)\}_{i=1,\dots,N}$ where $X$ is 
sampled uniformly over its $\mathrm{Range}(x)$. Based on the data a \ac{GMM} is fitted.
\begin{equation}
\begin{split}
X,Y \sim \mathcal{G}_{XY}(x,y|g_{XY})
\end{split}
\end{equation}

that represents the measurement curve as well as the superposed noise. If the 
\ac{GMM} accurately represents the measurement device one must have:
\begin{equation}
\begin{split}
& \Vert \mathbb{E}(Y|X)-f(x) \Vert_{L^2(\mathbb{R})}= \sqrt{\int_{\mathbb{R}}^{} \left(\mathbb{E}(Y|X)-f(x) \right)^2\mathrm{d}x}= \mathbb{E}(W) = 0\\
& \Vert \mathbb{V}(Y|X)-\sigma^2 \Vert_{L^2(\mathbb{R})} = \sqrt{\int_{\mathbb{R}} \left(\mathbb{V}(Y|X)-\sigma^2 \right)^2\mathrm{d}x} = 0.
\end{split}
\end{equation}

For the case:

\begin{equation}
\begin{split}
&f \colon [0,1] \to [0,0.8] \qquad x \mapsto x-0.2x^2\\
&\widetilde{f} \colon [0,1]\times(-\infty,+\infty) \to \mathbb{R} \quad x,w \mapsto y = x-0.2x^2+w\\
&W \sim \mathcal{N}(w|0,0.0025) \\
\end{split}
\end{equation}

a Gaussian mixture model was obtained using the $\mathrm{EM}$ algorithm. As stated \ac{BIC} or \ac{AIC} are 
commonly used to find the optimal number of components $K$.

$N=1000$ simulated data points and $K = 10$ were used to obtain the Gaussian mixture and:

\begin{equation}
\begin{split}
&\Vert \mathbb{E}(Y|X)-f(x) \Vert_{L^2[0,1]}= \sqrt{\int_{0}^{1} \left(\mathbb{E}(Y|X)-f(x) \right)^2\mathrm{d}x}=     5.24\times 10^{-7}\\
&\Vert \mathbb{V}(Y|X)-0.0025 \Vert_{L^2[0,1]}= \sqrt{\int_{0}^{1} \left(\mathbb{V}(Y|X)-0.0025 \right)^2\mathrm{d}x} = 6.20\times 10^{-6}.
\end{split}
\end{equation}

\begin{figure}[!ht]
    \centering
    \includegraphics[width=1.0\textwidth]{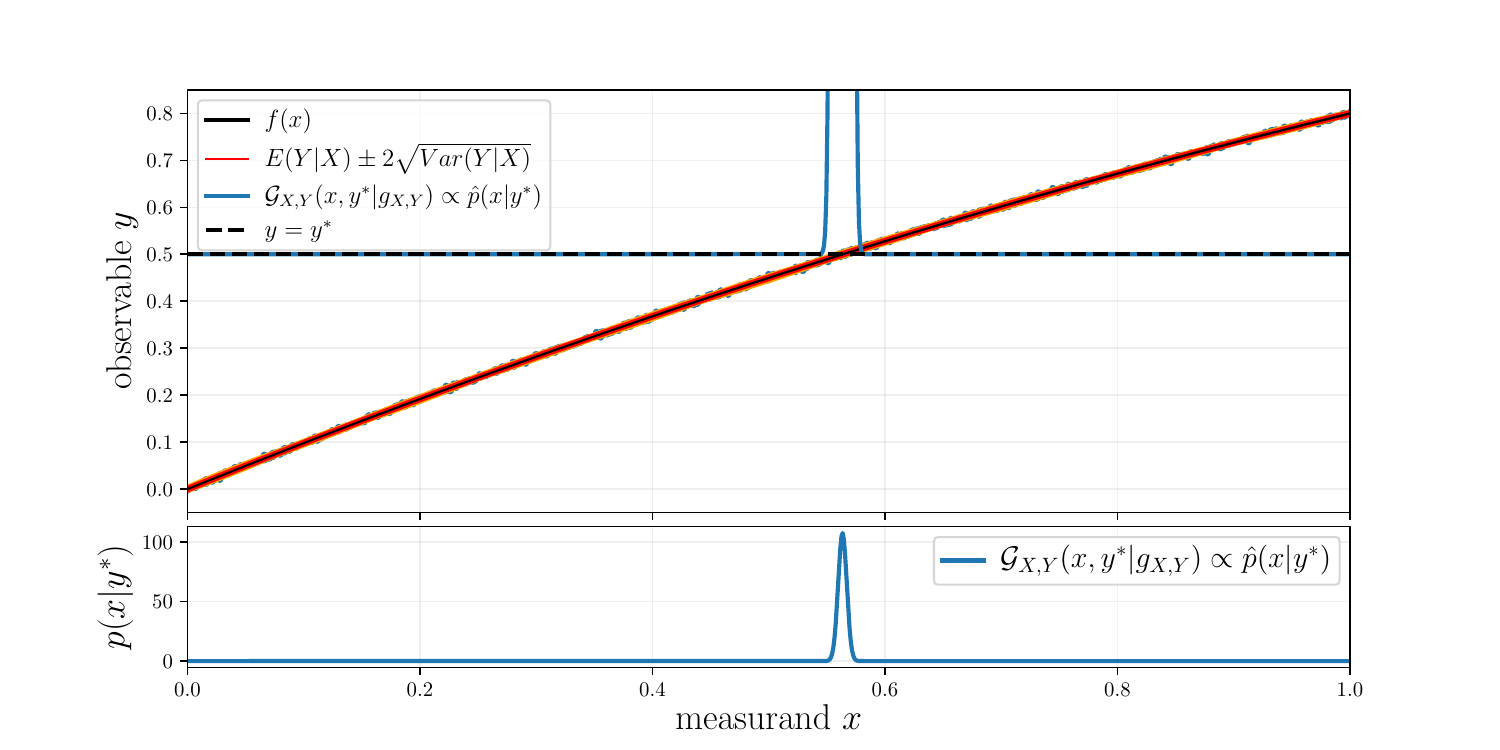}
    \caption{Modeling of a measurement system using \ac{GMM}; low noise case.}
    \label{fig:measurement_system_low}
\end{figure}

The metrics indicate that the obtained Gaussian mixture can accurately represent the measurement device; see \Cref{fig:measurement_system_low}.

\begin{figure}[!ht]
    \centering
    \includegraphics[width=1.0\textwidth]{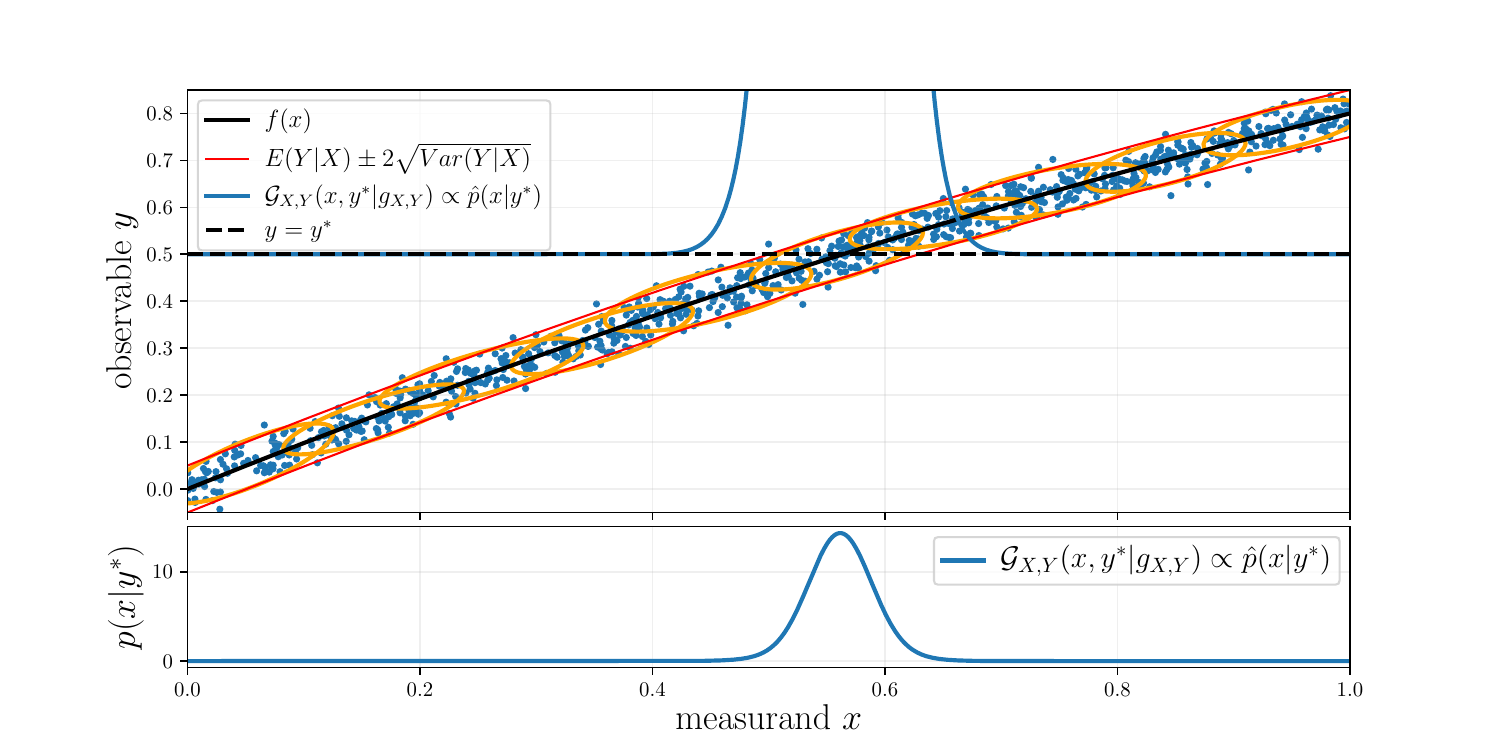}
    \caption{Modeling of a measurement system using \ac{GMM}; high noise case.}
    \label{fig:measurement_system_high}
\end{figure}

To be able to see the conditioning with the naked eye assume now the noise is distributed according
to $\mathcal{N}(w|0,0.03)$. The results
are shown in \Cref{fig:measurement_system_high}. The blue dots represent the
simulated measurement data, the radii of the orange ellipses
represent twice the square root of the eigenvalues of the covariance matrices of the Gaussian
components of the \ac{GMM}, the black curve represents the $f(x)$ and the red
curves represent $\mathbb{E}(Y|X) \pm 2\sqrt{\mathbb{V}(Y|X)}$. About $95\%$ 
of the data lie between the red curves. After
the observable is read of the measurement device as $y^*=0.5$ a \ac{PDF} in the form of a \ac{GMM} can be
assigned to the measurand $x$ using conditioning formulas found in \Cref{eq:conditioning_in_table} in \Cref{tab:gm_math} and alternatively in \Cref{prop:gmm_conditioning} in \Cref{sec:appendix}.
 
\section{Conclusion}

In this paper, we have presented a consistent and scalable probabilistic framework for the representation, propagation, and processing of measurement 
uncertainty in industrial and engineering systems, circular factories in particular. Departing from the traditional restriction of uncertainty representation to 
Gaussian distributions characterized solely by mean and variance, we have argued that many practically relevant measurement scenarios violate the 
assumptions underlying the Gaussian framework, particularly in the presence of nonlinear transformations, multimodal effects, constraints, and source uncertainty fusion.
To address these limitations, we proposed \acp{GMM} as a unifying and expressive representation for uncertain 
quantitative attributes. Owing to their universal approximation capability, finite-dimensional parameterization, and 
closure under a wide range of algebraic operations, \acp{GMM} enable uncertainty to be propagated and fused with high precision while remaining computationally tractable. We 
demonstrated that many operations of practical relevance--such as affine transformations, convolution, marginalization, conditioning, and Bayesian fusion--can be 
performed exactly and efficiently within the Gaussian mixture framework all in closed form. For cases where closed form propagation is not available, efficient ad hoc samplers
 combined with \ac{EM} based fitting provides an effective and scalable approximation strategy. Beyond the representation of uncertain quantities themselves, 
 we showed that entire measurement systems can be modeled probabilistically using \acp{GMM}. This enables a principled Bayesian treatment of measurement processes, where prior 
 knowledge and posterior distributions are all expressed within the same functional class. As a result, the complete measurement chain--from raw observables to 
 final measurement results--can be described by \ac{EM} fitting and algebraic operations on Gaussian mixture parameters, minimizing information loss and avoiding 
 multistage Gaussian approximations. Importantly, the framework remains compatible with established metrological practice through an explicit Gaussian fallback, 
 allowing results to be reduced to mean and standard deviation whenever required. The examples presented throughout the paper illustrate that 
 the proposed approach is not merely of theoretical interest but is well suited for real world applications, including tolerance analysis, 
 quality control, uncertainty propagation in multi stage systems, and the fusion of heterogeneous information sources. From a software and computational perspective, 
 the encapsulation of \ac{GMM} based operations allows users to benefit from improved uncertainty handling without increased conceptual or computational burden. 
 In summary, \ac{GMM} provide a robust, expressive, and computationally efficient foundation for modern uncertainty quantification in measurement science. By treating \acp{GMM} as a 
 native data type for uncertain quantities, the proposed framework enables more accurate uncertainty propagation, better risk assessment, and a 
 unified probabilistic treatment of aleatory and epistemic uncertainties by interpreting probabilitis as \ac{DOB}. Future work may focus on automated model selection, real time adaptive mixture refinement, and the integration of the framework into 
 standardized metrological software and digital twin architectures.

\section*{Acknowledgments}
Funded by:\\
- Project Name: SFB 1574, A Circular Factory for the Perpetual Product\\
- Funding Agency: Deutsche Forschungsgemeinschaft (DFG)\\
- Project ID: 471687386
\newpage
\appendix
\section{Appendix}
\label{sec:appendix}
Here we give a terse theorem like exposition blended with explanatory text to the 
basic results on \ac{GMM} that are used throughout the paper. 
We define the Gaussians, express the key operations on them, and then lift these operations
whenever possible to \acp{GMM}. The sequence in which the results are presented are based on mathematical construction 
and to avoid circularity which is simple terms is not to have logical flaws in the proofs even though the proofs are not included. 
As we do not present proofs but rather present the results it might be 
a bit more involved to convince yourself on the correctness of the order of presentation. We hope it enables the avid reader to build their 
personal library based of \acp{GMM} and make use of their powerful representation and reasoning in their respective field.
\subsection{Theorem Style Constructs}
We make heavy use of four theorem like constructs namely \textbf{Definitions}, \textbf{Propositions}, \textbf{Corollaries} and \textbf{Remarks}.
Definitions are used to formally define new concepts and objects. Propositions are used to state important properties or results that are not as general or 
significant as theorems and usually do not have names. Corollaries are used to state results that follow directly from a theorem or proposition. Remarks are used to 
provide additional insights, clarifications, notational conventions or observations related to the preceding content.
\subsection{Gaussians}
As the Gaussian \acp{PDF} are the atomic building blocks; it is of great importance to precisely define them and their parameters 
and the spaces that the said parameters belongs to. This will introduce a clear notation and from the computational and software perspective 
give a clear way to define the class of Gaussian \acp{PDF} if desired.
\begin{definition}[Gaussian \ac{PDF}]
  Let
\begin{equation}
  \begin{split}
    &d                     \in \mathbb{N}\setminus \{0,+\infty\}\\
    &x \in  \mathbb{R}^d,m \in \mathbb{R}^d\\
  &\Sigma \in \{A \in \mathbb{R}^{d \times d}\colon x^\top \boldsymbol{\cdot}A\boldsymbol{\cdot}x>0 \text{ for } x \neq 0\}
  \end{split}
\end{equation}
then the mapping
\begin{equation}
  \mathcal{N}_X(x|m,\Sigma) \coloneq \frac{\exp\left[-\frac{1}{2}(x-m)^\top \boldsymbol{\cdot}\Sigma^{-1}\boldsymbol{\cdot}(x-m)\right] }{\sqrt{\det{(2\pi\Sigma)}}}
\end{equation}
is called the Gauss function or the Gaussian \ac{PDF}.
\end{definition}

\noindent It is common in probability theory and statistics to have a  notation for ``When the \ac{RV} is distributed according to the \ac{PDF} $p(x)$'' and also 
as the Gaussian \ac{PDF} is so ubiquitous in the literature there exist a notation its  \ac{PDF} too which brings us to the following remark.
\begin{remark}[Notation]
  When the Random Variable $X$ is Gaussian distributed, we write $X \sim \mathcal{N}_X(x|m,\Sigma)$.
\end{remark}

\noindent Expectation value and the variance/covariance are two of the most important 
statistics of a \ac{PDF} and possibly the first two properties one would think of when facing a new \ac{PDF}. Expectation can be thought 
of as the average of the data that the \ac{PDF} represents and the variance can be thought of as the deviation of the data around their average. If 
one has a good grasp on how they are calculated in closed form in the case of Gaussians, they can be easily extended to \acp{GMM} as we will see later. 
Next you will find the familiar results in the case of Gaussians presented as propositions.

\begin{proposition}[Expectation of Gaussians]
\label{prop:appendix_gmm_expectation}
\begin{equation}
  \mathbb{E}(X \sim \mathcal{N}_X(x|m,\Sigma))=m
\end{equation}
\end{proposition}

\begin{proposition}[Covariance Matrix of Gaussians]
\label{prop:appendix_gmm_variance}
\begin{equation}
  \mathbb{V}(X \sim \mathcal{N}_X(x|m,\Sigma))=\Sigma
\end{equation}
\end{proposition}

\subsubsection{Marginalization}
\noindent In probability and statistics, \textbf{marginalization} is the process of obtaining 
the probability distribution of a subset of variables by summing or integrating 
over the other variables. Suppose we have a joint probability distribution for two 
random variables, $X$ and $Y$. To find the marginal distribution of $X$, we 
``marginalize out'' $Y$ by integrating over all possible values of $Y$. 
The result is the marginal probability of $X$ alone. Marginalization 
is fundamental in Bayesian inference, expectation calculations, 
and simplifying complex probabilistic models by focusing only on variables of interest. 
Thanks to the properties of the Gaussian distribution, marginalization can be performed 
in closed form and the result is also a Gaussian distribution.
\begin{proposition}[Gaussian Marginalization]
\label{prop:gaussian_marginalization}
Let $Z = \begin{bmatrix} X \\ Y \end{bmatrix} \in \mathbb{R}^d$, $Z \sim \mathcal{N}_Z(z|m_Z,\Sigma_Z)$
\begin{equation*}
  m_Z = \begin{bmatrix} {m_Z}_X \\ {m_Z}_Y \end{bmatrix},\quad
  \Sigma_Z = \begin{bmatrix}
{\Sigma_Z}_{XX} & {\Sigma_Z}_{XY} \\
{\Sigma_Z}_{YX} & {\Sigma_Z}_{YY},
\end{bmatrix}
\end{equation*}
then:
\begin{equation}
  \begin{split}
  &X    \sim   \mathcal{N}_X(x|m_X,\Sigma_X),\quad Y \sim   \mathcal{N}_Y(y|m_Y,\Sigma_Y) \\
  &{m_X}  = {m_Z}_X, \quad {m_Y}  = {m_Z}_Y, \quad {\Sigma_X}  = {\Sigma_Z}_{XX}, \quad {\Sigma_Y}  = {\Sigma_Z}_{YY}
  \end{split}
\end{equation}
\end{proposition}

\subsubsection{Conditioning}
\noindent In probability and statistics, \textbf{conditioning} in the case of \acp{PDF} refers 
to finding the conditional density of one continuous random variable given another. If $X$
 and $Y$ have a joint \ac{PDF}, the conditional \ac{PDF} of $X$ given $Y = y^*$ is 
 obtained by dividing the joint \ac{PDF} by the marginal \ac{PDF} of $Y$ at $y^*$. 
 This produces a new density that describes the distribution of $X$ when $Y$ is fixed at a 
 specific value. The conditional \ac{PDF} integrates to one over $X$ and reflects how 
 knowledge of $Y$ influences the likelihood of different values of $X$. Thanks to the properties of 
 the Gaussian distribution, conditioning can be performed in closed form and the result is also a Gaussian distribution.
\begin{proposition}[Gaussian Conditioning]
  \label{prop:gaussian_conditioning}
\begin{equation}\label{eq:gaussian_conditioning}
\begin{split}
  &X|Y    \sim   \mathcal{N}_{X|Y}(x|y,m_{X|Y},{\Sigma}_{X|Y})\\
  &{m_{X|Y}}  = m_X+\Sigma_{XY}\boldsymbol{\cdot}{\Sigma_{YY}}^{-1}\boldsymbol{\cdot}(y-m_Y)\\
  &{\Sigma_{X|Y}}  = \Sigma_{XX}-\Sigma_{XY}\boldsymbol{\cdot}{\Sigma_{YY}}^{-1}\boldsymbol{\cdot}\Sigma_{YX}\\
  &Y|X \sim   \mathcal{N}_{Y|X}(y|x,m_{Y|X},\Sigma_{Y|X}) \\
  &{m_{Y|X}}  = m_Y+\Sigma_{YX}\boldsymbol{\cdot}{\Sigma_{XX}}^{-1}\boldsymbol{\cdot}(x-m_X)\\
  &{\Sigma_{Y|X}}  = \Sigma_{YY}-\Sigma_{YX}\boldsymbol{\cdot}{\Sigma_{XX}}^{-1}\boldsymbol{\cdot}\Sigma_{XY}
  \end{split}
\end{equation}
\end{proposition}

\begin{proposition}[Convolution of Gaussians]
\label{prop:gaussian_convolution}
If $X \sim  \mathcal{N}_X(x|m_X,\Sigma_X)$ and $Y \sim  \mathcal{N}_Y(y|m_Y,\Sigma_Y)$ are stochastically independent, then $Z=X+Y \sim \mathcal{N}_Z(z|m_Z,\Sigma_Z) = \mathcal{N}_X(x|m_X,\Sigma_X) * \mathcal{N}_Y(y|m_Y,\Sigma_Y)$ where $*$ denotes the convolution operator and:
    \begin{equation}
        \begin{split}
          {m_Z}=& {m_X} + {m_Y}\\
          {\Sigma_Z}=& {\Sigma_X} + {\Sigma_Y}.\\
        \end{split}
      \end{equation}
\end{proposition}

\subsubsection{Fusion}
\noindent Two independent Gaussian \acp{PDF} about the same 
quantity can be combined into one through the Bayes' theorem. Each 
Gaussian may come from a sensor, model, or prior knowledge. 
The result is another Gaussian whose mean is a weighted average, giving more weight to the more 
reliable (lower uncertainty) source, and whose uncertainty is reduced. This is fundamental in sensor 
fusion, estimation, and tracking (e.g., Kalman filters). Its importance lies in providing an optimal 
way to merge information. Having a closed-form solution avoids numerical integration, 
making computation fast, stable, and scalable for real-time engineering systems.
\begin{proposition}[Bayesian Fusion of Gaussians]
  \label{prop:gaussian_bayesian_fusion}
Let $\mathcal{N}_X (x|m_s, \Sigma_s)$ denote a Gaussian density of $X$, then:
\begin{equation}
  \mathcal{N}_X (x|m_a, \Sigma_a) \boldsymbol{\cdot} \mathcal{N}_X (x|m_b, \Sigma_b) =  c_{ab} \mathcal{N}_X (x|m_{ab}, \Sigma_{ab})
\end{equation}
where:
\begin{equation}
  \begin{split}
    c_{ab}&= \frac{\exp\left[-\frac{1}{2}(m_a-m_b)^\top \boldsymbol{\cdot}(\Sigma_a+\Sigma_b)^{-1}\boldsymbol{\cdot}(m_a-m_b)\right] }{\sqrt{\det{(2\pi(\Sigma_a+\Sigma_b))}}}\\
    m_{ab}&=  (\Sigma_a^{-1}+\Sigma_b^{-1})^{-1} \boldsymbol{\cdot}(\Sigma_a^{-1}\boldsymbol{\cdot}m_a+\Sigma_b^{-1}\boldsymbol{\cdot}m_b)\\
    \Sigma_{ab}&= (\Sigma_a^{-1}+\Sigma_b^{-1})^{-1}
  \end{split}
\end{equation}
\end{proposition}

Knowing the normalization constant of the product of two Gaussians is important for several reasons. First, it allows us to 
compute the exact posterior distribution in Bayesian inference when both the prior and likelihood are Gaussian. 
This is crucial for applications like Kalman filtering, where we need to update our beliefs based on new measurements. 
Second, it provides a way to evaluate metric to compare different Gaussian distributions, which is useful in 
model selection and hypothesis testing.
\begin{corollary}[Normalization Constant of Bayesian Fusion of Gaussians]
  \begin{equation}
    \int_{\Omega} \mathcal{N}_X (x|m_a, \Sigma_a) \boldsymbol{\cdot} \mathcal{N}_X (x|m_b, \Sigma_b) \mathrm{d}x=  c_{ab}
  \end{equation}
\end{corollary}

\subsubsection{Sampling}
\noindent While sampling from a general \ac{PDF} and performing a Monte Carlo simulation is not a straight forward process; 
in the case of Gaussian \acp{PDF} it is easy to generate random numbers using only uniform random numbers, 
which are easy for computers to produce. By applying simple mathematical transformations, two 
independent standard Gaussian samples are obtained. This method is important because many engineering simulations, 
noise models, and Monte Carlo algorithms rely on Gaussian randomness. The Box-Muller transformation 
coupled with an affine transformation
provides an exact, closed-form way to convert samples drawn from a uniform distribution to samples drawn from the desired 
multivariate \ac{PDF}. Its simplicity, efficiency, and reliability makes Monte Carlo simulations and 
stochastic modeling with Gaussian \acp{PDF} feasible and widely used in engineering applications.

\begin{proposition}[Box-Muller Transformation]
Let $U$ and $V$ be independent samples drawn from the $\mathcal{U}(x|0,1)$ \ac{PDF}. Then 
for $R = \sqrt{-2 \log U},\Theta = 2\pi V$ the samples
$X = R \cos \Theta,Y = R \sin \Theta$ are independent samples drawn from the 
standard Gaussian \ac{PDF} $\mathcal{N}(x|0,1)$.
\end{proposition}

\begin{proposition}[Multivariate Gaussian Sampling]
\label{prop:multivariate_gaussian_sampling}
Let $Z \in \mathbb{R}^d$ such that $Z_i$ is a sample drawn 
from the \ac{PDF} $\mathcal{N}(z_i|0,1) \text{ for } i=1,\dots,d$. For the full 
rank matrix $A$ and the vector $m \in \mathbb{R}^d$, the sample $X = m + AZ$ 
will be a sample drawn from the \ac{PDF} $\mathcal{N}_X(x|m,AA^\top)$.
\end{proposition}

\section{\acp{GMM}}
As a reminder as seen in \Cref{def:gmm} 
The definition of \acp{GMM} is a direct extension of the definition of Gaussians, 
where we have a weighted sum of Gaussians instead of just 
one Gaussian. Now let us express the key operations on \acp{GMM}. 

\noindent Using the \Cref{prop:gaussian_marginalization} the \textbf{marginalization} of \acp{GMM} can be easily derived. The result is a 
\ac{GMM} with the same number of components as the original joint \ac{GMM},
\begin{proposition}[\ac{GMM} Marginalization]
\label{prop:gmm_marginalization}
Let $Z = \begin{bmatrix} X \\ Y \end{bmatrix} \in \mathbb{R}^d$, $Z \sim \mathcal{G}_Z(z|g_Z)$
\begin{equation*}
  {m_Z}_i = \begin{bmatrix} {{m_Z}_i}_X \\ {{{m_Z}}_i}_Y \end{bmatrix},\quad
  {\Sigma_Z}_i = \begin{bmatrix}
{{\Sigma_Z}_{i}}_{XX} & {{\Sigma_Z}_{i}}_{XY} \\
{{\Sigma_Y}_{i}}_{YX} & {{\Sigma_Z}_{i}}_{YY}
\end{bmatrix}
\end{equation*}
then:
\begin{equation}\label{eq:gmm_marginalization}
  \begin{split}
  &X    \sim   \mathcal{G}_X(x|g_X),\quad Y \sim   \mathcal{G}_Y(x|g_Y) \\
  &K_X = K_Y = K_Z\\
  &{\pi_X}_i = {\pi_Y}_i = {\pi_Z}_i\\
  &{m_X}_i  = {{m_Z}_i}_X, \quad {m_Y}_i  = {{m_Z}_i}_Y, \\
  &{\Sigma_X}_i  = {{\Sigma_Z}_i}_{XX}, \quad {\Sigma_Y}_i  = {{\Sigma_Z}_i}_{YY}
  \end{split}
\end{equation}
where the subscript $i$ denotes the $i$-th component of the resulting \ac{GMM}.
\end{proposition}

\noindent Using the \Cref{prop:gaussian_conditioning} and normalization of the weights the \textbf{conditioning} of \acp{GMM} can be easily derived. The result is a 
\ac{GMM} with the same number of components as the original joint \ac{GMM},
\begin{proposition}[\ac{GMM} Conditioning]
\label{prop:gmm_conditioning}
    \begin{equation}\label{eq:gmm_conditioning}
  \begin{split}
  &X|Y    \sim   \mathcal{G}_{X|Y}(x|y,g_{X|Y})\\
  & K_{X|Y}=K_Z\\
  &{m_{X|Y}}_i  = {m_i}_X+{\Sigma_{i}}_{XY}\boldsymbol{\cdot}{{\Sigma_{i}}_{YY}}^{-1}\boldsymbol{\cdot}(y-{m_i}_Y)\\
  &{\Sigma_{X|Y}}_i  = {\Sigma_{i}}_{XX}-{\Sigma_{i}}_{XY}\boldsymbol{\cdot}{{\Sigma_{i}}_{YY}}^{-1}\boldsymbol{\cdot}{\Sigma_{i}}_{YX}\\
  &{\pi_{X|Y}}_i =\frac{\pi_i \mathcal{N}(y|{m_Y}_i,{\Sigma_Y}_i)}{\sum_{j=1}^{K_{Y}}\pi_j \mathcal{N}(y|{m_Y}_j,{\Sigma_Y}_j)}\\
  &Y|X    \sim   \mathcal{G}_{Y|X}(y|x,g_{Y|X})\\
  & K_{Y|X}=K_Z\\
  &{m_{Y|X}}_i  = {m_i}_Y+{\Sigma_{i}}_{YX}\boldsymbol{\cdot}{{\Sigma_{i}}_{XX}}^{-1}\boldsymbol{\cdot}(x-{m_i}_x)\\
  &{\Sigma_{Y|X}}_i = {\Sigma_{i}}_{YY}-{\Sigma_{i}}_{YX}\boldsymbol{\cdot}{{\Sigma_{i}}_{XX}}^{-1}\boldsymbol{\cdot}{\Sigma_{i}}_{XY}\\
  &{\pi_{Y|X}}_i =\frac{\pi_i \mathcal{N}(x|{m_X}_i,{\Sigma_X}_i)}{\sum_{j=1}^{K_{X}}\pi_j \mathcal{N}(x|{m_X}_j,{\Sigma_X}_j)}\\
  \end{split}
\end{equation}
\end{proposition}

\noindent Using the \Cref{def:gmm} and the \Cref{prop:gaussian_bayesian_fusion} the \textbf{Bayesian fusion of \acp{GMM}} can be easily derived. 
If the first \ac{GMM} has $K_a$ components and the second \ac{GMM} has $K_b$ components, the result is a
\ac{GMM} with $K_a \boldsymbol{\cdot}K_b$ components, which is the result of pairing all the components of the two original \acp{GMM}.
\begin{proposition}[Bayesian Fusion for \acp{GMM}]
\label{prop:gmm_multiplication}
\begin{equation}
  \begin{split}
  &{\mathcal{G}_{X}}(x|g_a)=\sum_{i=1}^{K_a}{\pi_a}_i \mathcal{N}_X(x|{m_a}_i,{\Sigma_a}_i)\\
  &{\mathcal{G}_{X}}(x|g_b)=\sum_{j=1}^{K_b}{\pi_b}_j \mathcal{N}_X(x|{m_b}_j,{\Sigma_b}_j)
  \end{split}
\end{equation}
then The following holds:
\begin{equation}\label{eq:gmm_multiplication}
  \begin{split}
  &{\mathcal{G}_{X}}(x|g_a)\boldsymbol{\cdot}{\mathcal{G}_{X}}(x|g_b)=\sum_{i=1}^{K_a}\sum_{j=1}^{K_b}c_{a_i b_j}{\pi_a}_i{\pi_b}_j \mathcal{N}_X(x|{m_{a_ib_j}},{\Sigma_{a_ib_j}})\\
  &\int_{\Omega}{\mathcal{G}_{X}}(x|g_a)\boldsymbol{\cdot}{\mathcal{G}_{X}}(x|g_b)\mathrm{d}x=\sum_{i=1}^{K_a}\sum_{j=1}^{K_b}c_{a_i b_j}{\pi_a}_i{\pi_b}_j
  \end{split}
\end{equation}
where $c_{a_i b_j}$, ${m_{a_ib_j}}$ and ${\Sigma_{a_ib_j}}$ are as in \Cref{prop:gaussian_bayesian_fusion}.
This gives a closed form for the Bayesian fusion of two independent information contribution.
\end{proposition}

\subsection{Source mixture} 
If we have $J$ independent random generators (sources) for one and the same quantity $X$ and if each 
source is active with probability $w^j,$ with $w^j \geq 0, \sum_{j=1}^{J}w^j=1$; 
pooling all these sources leads to a common \ac{PDF} $p_X(x)$ that is the convex linear combination of the \acp{PDF} 
$p^j_X(x)$ of the sources; in summary:
\begin{equation}
  p_X(x)=\sum_{j=1}^{J}w^j p^{j}_{X}(x).
\end{equation}

In the case that the $p^j_X(x)$s are \acp{GMM}, the common \ac{PDF} $p_X(x)$ becomes a mixture of \acp{GMM}. It can be easily seen that $p_X(x)$ is then also 
a \ac{GMM}:

\begin{equation}
  \begin{split}
    p_X(x) =\sum_{j=1}^{J}w^j \mathcal{G}_{X}(x|g^j_X) =\sum_{j=1}^{J}w^j \sum_{i=1}^{K_j} \pi^j_i\mathcal{N}_{X}(x|m^j_i,\Sigma^j_i) =\sum_{j=1}^{J} \sum_{i=1}^{K_j} w^j \pi^j_i\mathcal{N}_{X}(x|m^j_i,\Sigma^j_i)\\
  \end{split}
\end{equation}

thus, $p_X(x)$ is a \ac{GMM} with $K = \sum_{j=1}^{J}K_j$ Gaussian components with weights 
$w^j\pi^j_i$. Note that the expectations $m^j_i$ and covariance matrices $\Sigma^j_i$ 
remain unmodified. This will lead to the following proposition.

\begin{proposition}[Source Mixture]
\label{prop:gmm_convex_hull}
    Let $X^j \sim \mathcal{G}_{X^j}(x|g^j_{X})$. The convex linear combination is also a \ac{GMM} and:
\begin{equation}
  \begin{split}
    X \sim \mathcal{G}_X(x) =\sum_{j=1}^{J}w^j \mathcal{G}_{X}(x|g^j_X) =\sum_{j=1}^{J}w^j \sum_{i=1}^{K_j} \pi^j_i\mathcal{N}_{X}(x|m^j_i,\Sigma^j_i) =\sum_{j=1}^{J} \sum_{i=1}^{K_j} w^j \pi^j_i\mathcal{N}_{X}(x|m^j_i,\Sigma^j_i)\\
  \end{split}
\end{equation}
\end{proposition}

Having an explicit formula might be a bit overwhelming, but it would enable the reader for a 
rather easy and clean computer implementation. The common \ac{PDF} can be written as:
\begin{equation}
  \begin{split}
    p_X(x) = \mathcal{G}_X(x|g_X)= \sum_{k=1}^{K}w_k \mathcal{N}_X(x|m_k,\Sigma_k).
  \end{split}
\end{equation}

The connection between the indices $(i,j)$ and $k$ can be established with the following bijection:
\begin{equation}
  \begin{split}
    (i,j) &\mapsto k \text{ with } k(i,j)=\begin{cases}
      \sum_{v=1}^{i \leq K_1} 1 \text{ for } j=1\\
      \sum_{u=1}^{j-1}K_u +\sum_{v=1}^{i \leq K_j} 1 \text{ for } j\geq 2
    \end{cases}\\
    i(j) &= 1,\ldots, K_j\\
    j &= 1,\ldots, J.
  \end{split}
\end{equation}
The weights $w_k$ are thus:
\begin{equation}
  w_k = w_{k(i,j)} = w^j \pi^j_i.
\end{equation}

Thus the implementation consists of concatenating all Gaussian components of 
the sources and multiplying their mixture weights by the corresponding source 
weights.

\begin{proposition}[Convolution of \acp{GMM}]\label{prop:gmm_convolution}
If $X \sim \mathcal{G}_X(x|g_X)$ and $Y \sim \mathcal{G}_Y(x|g_Y)$ are stochastically independent, then $Z = X + Y \sim \mathcal{G}_Z(z|g_Z) = \mathcal{G}_X(x|g_X) * \mathcal{G}_Y(y|g_Y)$ where $*$ denotes the convolution operator and:
    \begin{equation}\label{eq:gmm_convolution}
        \begin{split}
          &g_X       = (K_X,{\pi_X}_i,{m_X}_i,{\Sigma_X}_i)_{i=1,\dots,K_X}\\
          &g_Y       = (K_Y,{\pi_Y}_j,{m_Y}_j,{\Sigma_Y}_j)_{j=1,\dots,K_Y}\\
          &g_Z       = (K_Z,{\pi_Z}_{ij},{m_Z}_{ij},{\Sigma_Z}_{ij})_{\substack{i=1,\dots,K_X \\ j=1,\dots,K_Y}}\\
          &K_Z       = K_X \boldsymbol{\cdot} K_y\\
          &{\pi_Z}_{ij}       = {\pi_X}_i \boldsymbol{\cdot} {\pi_Y}_j\\
          &{m_Z}_{ij}= {m_X}_i + {m_Y}_j\\
          &{\Sigma_Z}_{ij}= {\Sigma_X}_i + {\Sigma_Y}_j.
        \end{split}
      \end{equation}
\end{proposition}

\noindent Bayesian fusion and convolution are two fundamental operations for combining information in the context of \acp{GMM}. 
They result in an increase in the number of components, which can lead to computational challenges. If one can find 
a closed form integral based metric for the difference between two \acp{GMM} as in \Cref{eq:l2_distance_gmm}, it would be possible to use 
optimization techniques to find a \ac{GMM} with a smaller number of components that approximates the result of the Bayesian fusion or convolution, thus mitigating the computational challenges.

\begin{proposition}[\ac{GMM} Sampling]
\label{prop:gmm_sampling}
To sample from a \ac{GMM}, one first selects a component according to the 
mixture weights, treating them as probabilities of a categorical distribution. Once a 
component is chosen, a sample is drawn from the corresponding Gaussian distribution 
defined by that component's mean and covariance. The makes sampling from a \ac{GMM} an exact process up to the exactness of the uniform samples 
that are used.
\end{proposition}
\bibliographystyle{plain}
\bibliography{references}
\end{document}